\documentclass[twoside]{article}

\usepackage[preprint]{aistats2027}

\usepackage[utf8]{inputenc}
\usepackage[T1]{fontenc}

\usepackage[
  backend=biber,
  style=authoryear,
  natbib=true,
  url=false,
  doi=true
]{biblatex}
\defbibheading{bibliography}[\refname]{\subsubsection*{#1}}

\usepackage{amsmath}
\usepackage{mathtools}
\usepackage{graphicx}
\usepackage{multirow}

\usepackage{url}
\usepackage{booktabs}
\usepackage{amsfonts}
\usepackage{amssymb}
\usepackage{dsfont}
\usepackage{nicefrac}
\usepackage{microtype}
\usepackage{xcolor}

\usepackage{twemojis}
\usepackage{etoc}
\usepackage{float}
\usepackage{placeins}
\usepackage{algorithm}
\usepackage{algpseudocode}
\usepackage[colorlinks=true]{hyperref}
\usepackage{cleveref}

\newcommand{\deq}{\coloneqq}
\newcommand{\T}{^\top}
\newcommand{\norm}[1]{\left\lVert#1\right\rVert}
\DeclareMathOperator{\tr}{tr}
\DeclareMathOperator{\Cov}{Cov}
\newcommand{\E}{\mathbb{E}}
\newcommand{\KL}[2]{D_{\mathrm{KL}}\big(#1 \,\|\, #2\big)}
\newcommand{\cG}{\mathcal{G}}
\newcommand{\cL}{\mathcal{L}}
\newcommand{\cN}{\mathcal{N}}
\newcommand{\R}{{\mathbb{R}}}
\newcommand{\Ptrue}[1][c]{P_{#1}}
\newcommand{\Qscaf}[1][c]{Q_{#1}^{0}}
\newcommand{\Qmean}[1][c]{Q_{#1}^{\phi}}

\newcommand{\mDiff}{\texttt{Diff}}
\newcommand{\mMean}{\texttt{M}}
\newcommand{\dCRBP}{\texttt{CR3BP}}
\newcommand{\dSynth}{\texttt{Synthetic}}
\newcommand{\dPlume}{\texttt{Smoke}}
\newcommand{\mScaffold}{\texttt{Scaffold}}
\newcommand{\mKO}{\texttt{KO}}
\newcommand{\mFullDiff}{\texttt{Full-Diff}}
\newcommand{\mRescale}{\texttt{Rescale}}
\newcommand{\mGainMix}{\texttt{Gain-Mix}}
\newcommand{\mGainDiff}{\texttt{Gain-Diff}}
\newcommand{\mExactGP}{\texttt{Exact-GP}}
\newcommand{\mExact}{\texttt{Exact}}

\usepackage{colortbl}
\usepackage{arydshln}

\usepackage{tikz}

\definecolor{tueRed}{RGB}{141,45,57}
\definecolor{tueDark}{RGB}{55,65,74}
\definecolor{tueDarkblue}{RGB}{65,90,140}
\definecolor{tueDarkgreen}{RGB}{50,110,30}
\definecolor{tueOrange}{RGB}{210,150,0}
\DeclareRobustCommand{\legline}[1]{%
  \tikz[baseline=-0.5ex]\draw[#1, line width=1.2pt] (0,0) -- (0.4,0);}
\DeclareRobustCommand{\legmarkercirclefull}[1]{%
  \tikz[baseline=-0.5ex]
    \node[circle, draw=#1, fill=#1, 
          inner sep=0pt, 
          line width = 1pt,
          minimum size=3pt] {};}
\newcommand*{\figlegendline}[1]{(\protect\tikz[baseline]{\protect\draw[#1, very thick] (0,.6ex) -- (1em,.6ex);})}
\definecolor{plumeGainMix}{HTML}{C8503C}
\definecolor{plumeGainDiff}{HTML}{8D2D39}
\definecolor{plumeFullDiff}{HTML}{0069AA}
\definecolor{plumeDiff}{HTML}{326E1E}

\hypersetup{linkcolor=tueDarkgreen,citecolor=tueDarkblue,urlcolor=tueOrange}

\begin{document}
\etocdepthtag.toc{mtoc}

\runningtitle{Learn Only Where You Can't Compute}
\runningauthor{Weber, Sagebiel and Hennig}

\twocolumn[
	\aistatstitle{Learn Only Where You Can't Compute: \\ Structured Corrections to Gaussian Process Posteriors}

	\aistatsauthor{ 
		Tobias Weber, Carla Sagebiel \and Philipp Hennig
	}	
	\aistatsaddress{ 
		Tübingen AI Center \\
		Tübingen, Germany
	}	
]

\begin{abstract}
	Physics-informed Gaussian processes (GPs) provide a principled framework for scientific inference by expressing uncertainty over empirical measurements, boundary conditions, and governing equations.
	However, they struggle to scale to larger datasets and nonlinear information.
	Conversely, diffusion models excel at learning complex distributions over functions from massive datasets, but struggle in scientific contexts where training data are scarce and physical or observational information must be incorporated.
	We propose a structured posterior-correction framework that combines a tractable Gaussian reference with data-driven corrections via generative learning. 
	After a mean correction and covariance calibration, a diffusion model learns a targeted correction in a low-dimensional subspace selected to capture discrepancies from the reference, while the complement retains its Gaussian law.
	Across three scientific inference tasks, these corrections recover hidden modes, improve sample structure, and reduce prediction error relative to the Gaussian reference. Comparisons with full-field generative models reveal task-dependent benefits and tradeoffs between predictive accuracy and computational cost.
\end{abstract}

\section{Introduction}
\label{sec:introduction}

Scientific inference often combines heterogeneous sources of information about an unknown function or state, such as noisy measurements, boundary conditions, physical laws, numerical models, and historical data.
These sources typically leave the quantity of interest only partially determined, making uncertainty quantification an essential part of the inference problem~\citep{stuart2010inverse,hennig2022probnum}.
A central challenge is therefore not only to predict a function, but to construct a conditional distribution that combines the available information while remaining computationally tractable.

\begin{figure*}[t]
	\centering
	\includegraphics[width=\textwidth]{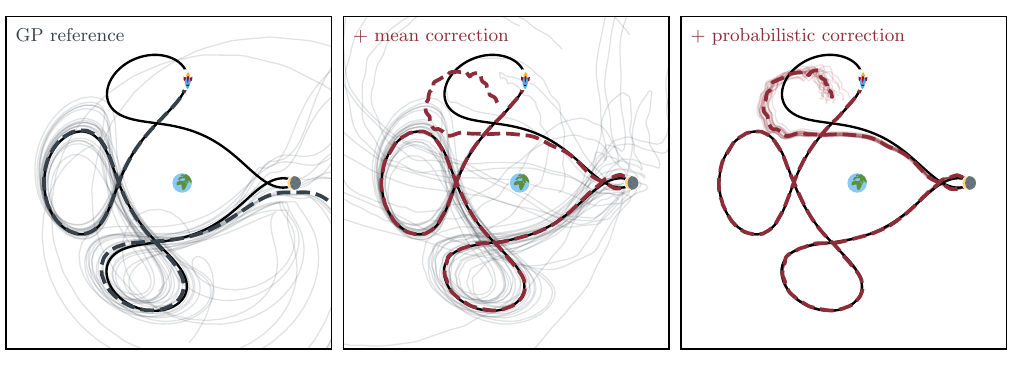}
	\caption{Illustrative correction pipeline for a \texttt{CR3BP} trajectory in the rotating Earth--Moon frame: target orbit (\legline{black}), Gaussian reference (\legline{tueDark,dashed}), and learned corrections (\legline{tueRed,dashed}); thin curves are samples.
		\textit{Left:} reference mean and samples.
		\textit{Middle:} mean correction shifts the reference toward the target.
		\textit{Right:} covariance and diffusion correction refines the ensemble spread.
		Earth $\texttwemoji{globe showing Europe-Africa}$ and Moon $\texttwemoji{waning crescent moon}$ mark their normalized locations; $\texttwemoji{rocket}$ marks the trajectory start.}
	\label{fig:cr3bp_pipeline}
\end{figure*}

Probabilistic function-space methods provide a natural way to combine such information.
Gaussian processes (GPs), in particular, place a distribution directly over the unknown function and admit analytic conditioning on linear information such as measurements, boundary values, or derivative observations~\citep{rasmussen2005gaussian,raissi2017machine,pfoertner2024physics}.
Probabilistic numerical methods extend this perspective to information supplied by differential equations and numerical computations~\citep{hennig2022probnum,schmidt2021probabilistic,kramer2022probabilistic}.
The resulting Gaussian posterior can encode substantial prior, physical, and observational structure before any learning from data is required.
It provides both a predictive mean and a structured representation of the uncertainty that remains.

Such a probabilistic model is nevertheless only as expressive as the assumptions from which it is constructed.
Incomplete physics, unresolved scales, latent nonlinear effects, or model mismatch may cause the true conditional law to differ systematically from its Gaussian approximation.
Some of this discrepancy amounts to an incorrect location or spread; other effects produce multimodality or dependencies that a Gaussian cannot represent.
Expressive generative models offer a complementary route. Diffusion models, in particular, can learn rich distributions over fields and trajectories~\citep{ho2020denoising,song2021score,cachay2023dyffusion,rozet2023score}.
Learning the entire conditional law, however, also asks the generative model to represent structure already available from the Gaussian posterior.
When scientific training data are expensive to obtain, there is a particular incentive to make use of this information.
This raises a question: \emph{how much of a scientific conditional distribution needs to be learned when a useful Gaussian approximation can already be computed?}

We turn this idea into a structured posterior-correction framework, illustrated in~\Cref{fig:cr3bp_pipeline}.
An analytically conditioned Gaussian serves as a probabilistic \emph{scaffold} (\Cref{fig:cr3bp_pipeline}, left).
A mean correction adjusts its predictive center across the whole field (middle).
The subsequent probabilistic correction recalibrates its covariance and fits a mixture or diffusion model in a selected low-dimensional subspace (right).
This learned law captures remaining discrepancies in conditional spread and non-Gaussian shape, while the complement retains its Gaussian law.

Our contributions are threefold.
First, we formulate a correction framework that uses an available Gaussian posterior to reduce the part of the conditional distribution assigned to a generative model.
Gaussian and subspace KL decompositions explain which discrepancies the corrections can repair and what remains outside the learned block.
Second, we develop a data-driven subspace search based on estimated validation log-score improvement, making the value of a richer law the criterion for where to learn it.
Third, we evaluate this restricted learning task on regression and smoke transport problems with tractable posterior references and on numerical trajectory reconstruction, using learning curves and measured run times alongside distributional scores.
Small learned blocks improve on the calibrated scaffold and can improve the structure of individual draws even when marginal scores are similar.
The experiments also show where full-field models remain preferable, while the restricted model still offers a useful accuracy--cost tradeoff.

\section{Method}
\label{sec:method}

We model a conditional distribution $\Ptrue$ by correcting an available Gaussian approximation $\Qscaf$.
The scaffold $\Qscaf{}$ provides an initial predictive mean and uncertainty geometry that we refine by a mean correction, a pooled covariance calibration, and a flexible law on a selected low-dimensional block, while retaining the Gaussian law on the complement.
The following decomposition motivates this sequence and makes its limitations explicit.

\textbf{Gaussian scaffold and correction objective.}
Let $f \in \R^d$ denote a discretized function of interest, and let $c$ represent the available information about $f$, such as observations, boundary conditions, or mechanistic inputs.
Given training pairs $(f^{(n)}, c^{(n)})$, our goal is to approximate the unknown conditional law $\Ptrue \deq \cL(f \mid c)$.
We assume that part of this inference problem can already be solved analytically via a tractable Gaussian conditioning step, yielding a reference distribution $\Qscaf = \cN(m_c, C_c)$.
The construction of this scaffold is application dependent.
In particular, probabilistic numerical solvers \citep{kramer2022probabilistic, weiland2025gmrf} can construct Gaussian conditional distributions that incorporate mechanistic information such as differential equations, boundary conditions, and observations \citep{pfoertner2024physics}.

To quantify the discrepancy between the true conditional law $\Ptrue$ and the Gaussian scaffold $\Qscaf$, we consider the Kullback--Leibler divergence, which arises from the excess risk of the scaffold when choosing the logarithmic score.
Writing $\mu_c = \E[f \mid c]$ and $\Sigma_c = \Cov(f \mid c)$, this discrepancy decomposes exactly as
\begin{align}
	\KL{\Ptrue}{\Qscaf}
	 & = \underbrace{\tfrac12 \norm{\mu_c - m_c}^2_{C_c^{-1}}}_{\text{location mismatch}} \nonumber                                         \\
	 & \; + \underbrace{\tfrac12 \big[\tr(C_c^{-1} \Sigma_c) - \log\det(C_c^{-1} \Sigma_c) - d\big]}_{\text{covariance mismatch}} \nonumber \\
	 & \; + \underbrace{\KL{\Ptrue}{\cN(\mu_c, \Sigma_c)}}_{\text{non-Gaussian structure}}.
	\label{eq:kl_decomposition}
\end{align}
This identity assumes a density with finite second moments, finite divergences, and nonsingular $C_c$ and $\Sigma_c$.
For numerically rank-deficient scaffolds we work in resolved coordinates with a positive covariance floor (Appendix~\ref{app:method:rescale}).
A shift in the scaffold mean cannot repair its spread or shape, but it can remove the location mismatch (cf. Eq.~\eqref{eq:kl_decomposition}, Term 1) across the entire field.
This motivates learning a full-field mean correction before deciding where a richer probability law is needed.

\textbf{Conditional mean correction.}
At the population level, the mean discrepancy $\mu_c-m_c$ minimizes the precision-weighted regression loss.
Its empirical counterpart is
\begin{equation}
	\frac{1}{N} \sum_{n=1}^{N} \norm{f^{(n)} - m_{c^{(n)}} - \delta_\phi(c^{(n)})}^2_{C_{c^{(n)}}^{-1}}.
\end{equation}
The precision $C_{c^{n}}^{-1}$ weights errors according to the scaffold uncertainty.
Unweighted squared error has the same unrestricted population minimizer and can be used alternatively.
We use then the fitted shift $\delta_\phi(c)$ to define the corrected mean $\hat\mu_c = m_c + \delta_\phi(c)$, residual $\rho_\phi=f-\hat\mu_c$, and the mean-corrected scaffold $\Qmean = \mathcal{N}(\hat\mu_c, C_c)$.

\textbf{Covariance mismatch correction.}
After correcting the mean, we target the covariance mismatch in Term~2 of~\eqref{eq:kl_decomposition}.
We first whiten the residuals by the scaffold covariance, so that the scaffold predicts unit variance in every direction, and estimate their covariance pooled across training contexts:
\begin{equation}
	z = C_c^{-1/2}(f - \hat\mu_c),
	\qquad
	\widehat{\Cov}(z)
	= R\operatorname{diag}(\hat\lambda_i)R^\top,
\end{equation}
Here, $r_i$ is the $i$th column of $R$, and $\hat\lambda_i$ is the estimated residual variance along that direction.
Values above or below one indicate that the scaffold underestimates or overestimates this variance, respectively.

To decide which variance corrections are supported by the data, we measure their potential log-score gain.
For a centered Gaussian coordinate, replacing unit variance by $\hat\lambda_i$ gives
\begin{equation}
	\kappa_i = \tfrac12
	\bigl(\hat \lambda_i - \log\hat \lambda_i - 1 \bigr).
	\label{eq:kappa}
\end{equation}
We set $\tilde\lambda_i=\hat\lambda_i$ when $\kappa_i$ exceeds a sampling-noise threshold, and retain the scaffold variance, $\tilde\lambda_i=1$, otherwise.
We then map these calibrated variances back to the original field coordinates using
\begin{equation}
	L_c = C_c^{1/2}R \operatorname{diag}(\tilde \lambda_i^{1/2}),
	\label{eq:cholesky_factor_calibration}
\end{equation}
where $C_c^{1/2}$ denotes the chosen square-root factor, the calibrated reference and its standardized coordinates are
\begin{equation}
	Q_c^\mathrm{G} = \cN(\hat\mu_c, L_cL_c^\top),
	\qquad
	u = L_c^{-1}(f - \hat\mu_c).
\end{equation}
Thus $u$ is standard normal under the calibrated reference $Q_c^\mathrm{G}$.
Because calibration pools residuals across contexts, it can leave context-dependent covariance mismatch and fitted mean error.
The next stage addresses these remaining discrepancies together with non-Gaussian structure.

\textbf{Probabilistic correction objective.}
To model this discrepancy without replacing the Gaussian in all $d$ directions, we assume an orthonormal matrix $U \in \R^{d \times k}$ and learn only a conditional law $q_\theta(\cdot \mid c)$ for $w = U^\top u$.
Writing $U_\perp$ for its orthogonal complement, the corrected density is
\begin{equation}
	q(u \mid c)
	= q_\theta(U^\top u \mid c)\,
	\cN(U_\perp^\top u; 0, I_{d-k}).
	\label{eq:family}
\end{equation}
For $w_\perp=U_\perp^\top u$, its divergence from the true residual law decomposes as
\begin{align}
	\KL{P_{u\mid c}}{q(\cdot\mid c)}
	 & =\KL{P_{w\mid c}}{q_\theta(\cdot\mid c)} \nonumber     \\
	 & \quad+\KL{P_{w_\perp\mid c}}{\cN(0,I_{d-k})} \nonumber \\
	 & \quad+I(w;w_\perp\mid c).
	\label{eq:kl_probabilistic_decomposition}
\end{align}
For fixed $U$, only the first term depends on $\theta$ and we fit the block law either by maximizing a Gaussian mixture likelihood~\citep{bishop1994mixture} or by training a diffusion model with a denoising objective~\citep{ho2020denoising}; the Gaussian complement requires no learned density.
However, even an exact block law leaves the complement mismatch and the dependence between block and complement.
Whitening alone does not remove these terms.

The remaining question is where this block should be placed.
Relative to the calibrated Gaussian, replacing the law of $w$ yields an improvement
\begin{align}
	\cG(U,q_\theta & ) \nonumber      \\
	=              & \E_{c,u}\!\left[
		\log q_\theta(U^\top u\mid c)
		-\log\cN(U^\top u;0,I_k)
		\right].
	\label{eq:gain}
\end{align}
For a fixed $U$, maximizing this gain over $q_\theta$ is equivalent to fitting the projected conditional law $P_{w \mid c}$ with the largest attainable gain being
\begin{equation}
	\E_c\!\left[
		\KL{P_{w\mid c}}{\cN(0, I_k)}
		\right].
\end{equation}
A positive population gain $\cG$ guarantees a KL improvement for the given block and fitted law.
Choosing $k\ll d$ reduces the stochastic dimension of the learned correction, while the calibrated Gaussian supplies the law on the complement.

\textbf{Subspace selection.}
Choosing $U$ is the difficult part.
Fitting a separate diffusion model for every candidate subspace is computationally prohibitive, so we estimate~\eqref{eq:gain} with a lightweight density model and train the final diffusion model only after selecting the block.
We use a product of one-dimensional Gaussian mixtures, conditioned on $c$ when needed, whose exact likelihood provides a held-out gain for each projected direction.

Log-score gain can favor sharp discrepancies with negligible amplitude in the original
function space.
Before evaluating it, we therefore add matched isotropic Gaussian noise of standard deviation $\tau$ to both the residual law and the Gaussian reference and repeat the covariance calibration.
This creates a soft noise floor, measured in the units of $f$, that decreases discrepancies smaller than $\tau$; setting $\tau=0$ recovers the original objective.

We initialize the search directions from residual-PCA, independent-component and excess-variance ($\hat\lambda$-span) proposals (\Cref{app:method:selection}).
We score these candidate blocks, refine the most promising frame together with its mixture density by gradient ascent on the training gain $\cG$, retaining the checkpoint with the highest validation gain.
Finally, we select directions whose validation gains exceed $\max\{g_\mathrm{min},p/n_{\mathrm{val}}\}$, with minimum gain threshold $g_\mathrm{min}=0.01$ and $p$ the parameter count of a one-dimensional mixture.
This is a practical threshold, not a statistical guarantee accounting for the adaptive search.

\textbf{Sampling from the corrected posterior.}
For $k>0$, we map the selected directions from the smoothed selection coordinates back to the original standardized coordinates $u$ and fit the final law $q_\theta$ to $w=U^\top u$.
A mixture from the selection family can be refitted as $q_\theta$; alternatively, we train a conditional diffusion model by denoising score matching~\citep{ho2020denoising}, allowing the final law to capture joint dependence between the selected coordinates.
A corrected sample is then
\begin{align}
	f^\star
	  & =\hat\mu_c + L_c(Uw + U_\perp \epsilon), \nonumber \\
	w & \sim q_\theta(\cdot \mid c),
	\qquad
	\epsilon \sim \cN(0, I_{d-k}),
	\qquad w \perp \epsilon.
	\label{eq:posterior_sampling}
\end{align}
The corrected mean $\hat\mu_c$ sets the center, $L_c$ from equation~\eqref{eq:cholesky_factor_calibration} combines the scaffold geometry with its covariance calibration, and $q_\theta$ replaces the Gaussian law only in the selected span.
For $k=0$, this reduces to $Q_c^\mathrm{G}$, for $k=d$, the family becomes a full residual model.
The experimental full-residual baselines also differ in calibration and architecture, as described in Appendix~\ref{app:method:laws}.

\section{Related work} \label{sec:related-work}

\textbf{Gaussian conditioning and probabilistic numerics.}
Gaussian processes provide closed-form inference under linear-Gaussian information, including observations and linear physical constraints~\citep{rasmussen2005gaussian,raissi2017machine,pfoertner2024physics}.
Probabilistic numerical methods extend this perspective to differential equations and numerical computation~\citep{hennig2022probnum,schmidt2021probabilistic,kramer2022probabilistic}, including sparse Gauss Markov formulations for PDE solutions~\citep{rue2005gaussian,weiland2025gmrf}.
For Gaussian measures conditioned on nonlinear observations of finitely many linear functionals, \citet{chen2024gaussian} derive a decomposition into a Gaussian component and a finite-dimensional non-Gaussian component.
In our setting, the discrepancy from an available Gaussian approximation is learned from data, including the directions in which a richer law is fitted.

\textbf{Low-dimensional corrections of reference distributions.}
The idea that a high-dimensional target can differ from a reference in a small subspace underlies non-Gaussian component analysis~\citep{blanchard2006search}, likelihood-informed subspaces~\citep{cui2014likelihood}, certified posterior dimension reduction~\citep{zahm2022certified}, and lazy transport maps~\citep{brennan2020lazy}.
Lazy maps relate restricted density-ratio corrections to projected KL gain and identify correction subspaces through gradient-based diagnostics.
Score-ratio matching learns such diagnostics from samples, including for conditional generative modeling~\citep{baptista2025score}.
We develop a correction procedure for scientific Gaussian posteriors that combines mean correction, covariance calibration, and data-driven selection of a residual block.
A lightweight conditional mixture estimates validation log-score gain, directing flexible modeling towards residual discrepancies for which a richer law improves on the calibrated Gaussian.
A mixture or diffusion model then learns the block law, while the calibrated Gaussian supplies the law on the complement.

\textbf{Model discrepancy and residual correction.}
Bayesian calibration represents the difference between a simulator and observations through a discrepancy process~\citep{kennedy2001bayesian,brynjarsdottir2014learning,plumlee2017bayesian}.
Learned residual models provide a more flexible alternative. For example, CorrDiff predicts a conditional mean and uses diffusion for the remaining variability for downscaling climate models~\citep{mardani2025residual}, while score-based approaches learn conditional distributions for data assimilation and PDE inference~\citep{rozet2023score,huang2024diffusionpde}.
We similarly learn residual structure, but start from a probabilistic scaffold and retain a calibrated Gaussian complement.
Our experiments examine what this restriction preserves and where full-field models remain preferable.

\raggedbottom
\section{Experiments}
\label{sec:experiments}

We examine how much of a conditional distribution can be recovered by learning a correction to an available Gaussian model.
Experiment 1 (\texttt{Synthetic}) tests the recovery of known hidden modes as training data increase.
Experiment 2 (\texttt{Smoke}) challenges the correction with measurement-dependent uncertainty over distinct smoke directions and examines the accuracy--cost tradeoff.
Experiment 3 (\texttt{CR3BP}) tests whether the correction improves a probabilistic numerical solver.
Across these problems, we assess predictive accuracy and the structure of individual samples.

\textbf{Methods.}
Each task supplies its own Gaussian scaffold (\mScaffold) and conditioning information.
\mScaffold{}+\mMean{} applies the mean correction, while \mRescale{} additionally recalibrates the covariance.
The selected subspace methods leverage a product mixture (\mGainMix{}) or diffusion (\mGainDiff{}) model; \mFullDiff{} learns the full residual law.
We compare against conditional diffusion directly on the field (\mDiff{}) and a Kennedy--O'Hagan GP discrepancy (\mKO{})~\citep{kennedy2001bayesian}, which adds a Gaussian process to the reference model to learn systematic differences between its predictions and the observed data.
In both \texttt{Synthetic} and \texttt{Smoke} we have access to the exact posterior (\mExact) and its moment-matched Gaussian (\mExactGP) as references.
For both \mFullDiff{} and \mDiff{} we use U-Nets~\citep{ronneberger2015unet} on \texttt{Smoke} and MLPs on \texttt{Synthetic} and \texttt{CR3BP}; all coefficient laws use MLPs.
These comparisons reflect differences in architecture and representation as well as the subspace restriction.
Each method underwent hyperparameter sweeps during development, after which its configuration was fixed for the reported experiments.
Details are provided in the corresponding appendices.

\textbf{Evaluation.}
The main comparisons use three seeds, $500$ test contexts, and  $100$ draws per context and method.
CRPS and Cram\'er distance $d_\mathrm{C}$ assess predictive distributions at individual coordinates, while energy score (ES) assesses complete fields.
Coverage $C_{90}$ checks whether central 90\% intervals contain the reference values at the expected frequency.
Accurate marginal predictions need not imply coherent joint samples.
We therefore define a physical consistency distance (PCD), which assigns each sample a negative log-density score under a smoothed exact posterior and compares the distributions of these scores for generated and exact draws.
Lower PCD indicates closer sample agreement;
Appendix~\ref{app:setup:metrics} gives the construction.
This diagnostic complements predictive scores but does not fully characterize the distribution.
\Cref{tab:main} summarizes the results aggregated across three seeds; metric definitions, seed variation, and additional results appear in the corresponding appendices.

\begin{table*}[t]
\centering\normalsize
\setlength{\tabcolsep}{2pt}
\caption{Main results, mean over three seeds. \dSynth{}: $d{=}512$, $n{=}2000$; \dPlume{}: $d{=}1024$, $n{=}8192$; \dCRBP{}: $d{=}2004$, $n{=}7500$. Lower is better except $C_{90}$ (target $0.90$). Baselines are above the dashed line, scaffold corrections below it, and exact references below the solid rule (not available for \dCRBP{}). Bold: mean $\pm$ one s.d.\ overlaps the best non-reference method.}
\label{tab:main}
\begin{tabular*}{\textwidth}{@{\extracolsep{\fill}}lrrrrrrrrrrrr@{}}
\toprule
& \multicolumn{4}{c}{\dSynth{}} & \multicolumn{4}{c}{\dPlume{}} & \multicolumn{4}{c}{\dCRBP{}} \\
\cmidrule(lr){2-5}\cmidrule(lr){6-9}\cmidrule(l){10-13}
Method & CRPS & $d_{\mathrm C}$ & $C_{90}$ & PCD & CRPS & $d_{\mathrm C}$ & $C_{90}$ & PCD & CRPS & ES & $C_{90}$ & Jacobi \\
\midrule
\mScaffold{} & 0.2859 & 0.1884 & 0.30 & 667 & 0.0714 & 0.0585 & 0.92 & 7072 & 0.1588 & 11.26 & \textbf{0.89} & 1.46 \\
\mKO{} & 0.2130 & 0.0427 & \textbf{0.90} & 1242 & 0.0714 & 0.0584 & 0.92 & 7079 & 0.2726 & 19.03 & 0.96 & 5.50 \\
\mDiff{} & 0.2208 & 0.0575 & 0.86 & 1256 & \textbf{0.0434} & \textbf{0.0014} & 0.89 & \textbf{37} & 0.1785 & 10.91 & 0.99 & 2.89 \\
\addlinespace[2pt]
\hdashline[0.4pt/2pt]
\addlinespace[2pt]
\mScaffold{}+\mMean{} & 0.2744 & 0.1659 & 0.33 & 512 & 0.0619 & 0.0385 & 0.94 & 6906 & 0.1178 & 7.83 & 0.93 & 2.01 \\
\mRescale{} & 0.2014 & 0.0193 & 0.94 & 406 & 0.0595 & 0.0336 & 0.92 & 5190 & 0.0916 & \textbf{7.60} & 0.94 & 0.79 \\
\mGainMix{} & \textbf{0.1936} & \textbf{0.0026} & 0.89 & \textbf{23} & 0.0473 & 0.0088 & 0.91 & 1022 & \textbf{0.0844} & \textbf{7.24} & 0.96 & 0.68 \\
\mGainDiff{} & \textbf{0.1934} & \textbf{0.0027} & 0.89 & 32 & 0.0457 & 0.0051 & \textbf{0.90} & 154 & \textbf{0.0815} & \textbf{7.06} & 0.93 & \textbf{0.44} \\
\mFullDiff{} & 0.1947 & 0.0050 & 0.87 & 958 & \textbf{0.0450} & 0.0049 & 0.88 & 249 & 0.0902 & 7.65 & 0.97 & 1.26 \\
\midrule
\mExactGP{} & 0.2007 & 0.0175 & 0.95 & 407 & 0.0459 & 0.0062 & 0.91 & 1256 & -- & -- & -- & -- \\
\mExact{} & 0.1921 & 0.0000 & 0.90 & 0 & 0.0429 & 0.0000 & 0.90 & 0 & -- & -- & -- & -- \\
\bottomrule
\end{tabular*}
\end{table*}

\subsection{\texttt{Synthetic}}
\label{subsec:exp_synthetic}

On a periodic grid, the target is $f=g+h+Bz$, where $g$ is a coarse GP, $h$ an independent fine-scale GP, and the columns of $B$ are smooth bumps between adjacent sensors, with signs $z\in\{-1,+1\}^{16}$ drawn uniformly.
Sixteen sensors observe $y=Af+\varepsilon$, which we provide as conditioning information $c$.
Since $AB=0$, the measurements do not identify the signs, and the exact posterior is
\begin{equation}
	p(f\mid y)=2^{-16}\textstyle\sum_z\cN\bigl(m_G(y)+Bz,C_G\bigr),
\end{equation}
where $(m_G,C_G)$ are the posterior moments of $g+h$ under the true sensor noise.
The scaffold conditions $g$ alone using one averaged noise level.
It therefore misses both Gaussian structure and the bump modes (Appendix~\ref{app:msgp}).

\begin{figure}[!t]
	\centering
	\includegraphics[width=\linewidth]{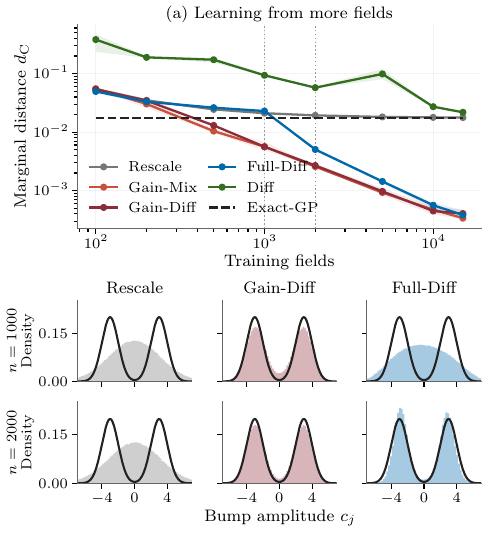}
	\caption{\dSynth{}. \textit{Top:} Cram\'er distance $d_{\mathrm C}$ to the exact posterior versus training-set size; mean and range over three seeds. \textit{Bottom:} pooled bump amplitudes at $n=1000$ and $2000$ (dotted lines; single seed) against the exact density (black). \mRescale{} cannot split the sign modes; \mGainDiff{} does so by $n=1000$, \mFullDiff{} only by $n=2000$.}
	\label{fig:synthetic_curve}
\end{figure}

The \texttt{Synthetic} columns of \Cref{tab:main} report results at $n_\mathrm{train}=2000$, while~\Cref{fig:synthetic_curve} shows Cram\'er distance across number of training pairs $(f^{(n)}, c^{(n)})$.
Gaussian calibration (\mRescale{}) recovers much of the missing spread, raising coverage $C_{90}$ from $0.30$ to $0.94$, but leaves the two sign modes at each bump unresolved.
The GP discrepancy baseline (\mKO{}) achieves nominal coverage, but its Cram\'er distance remains higher than that of $\mRescale{}$.
Learning a distribution on the subspace resolves this discrepancy. From $n_\mathrm{train} = 500$, both \mGainMix{} and \mGainDiff{} improve beyond the moment-matched Gaussian \mExactGP{} in marginal distance (\Cref{fig:synthetic_curve}).
The selector retains exactly $16$ coordinates on every seed, matching the number of hidden bumps.
A product mixture is particularly well suited to this construction, whose non-Gaussian structure arises from $16$ independent binary signs.
Therefore, $\mGainMix{}$ achieves marginal accuracy comparable to $\mGainDiff{}$.

The bump-amplitude distributions estimated from posterior samples explain the learning curves.
In the illustrated seed, \mGainDiff{} already separates the modes at $n=1000$, while \mFullDiff{} does so at $n=2000$.
These results suggest that restricting the correction to a subspace containing the non-Gaussian structure provides a useful inductive bias, enabling the model to recover the missing modes with fewer training examples in this experiment.

\subsection{\texttt{Smoke}}
\label{subsec:exp_plume}

We next consider smoke emitted by a chimney on a $32\times32$ grid under an uncertain wind direction $\varphi$.
Nine noisy sensors measure the smoke concentration $f$.
Some readings identify the wind direction sharply, while others leave several directions plausible.
We examine whether a low-dimensional correction can capture this measurement-dependent uncertainty.
The concentration is $f = b + Qg_\varphi$, where $b$ is a small GP background, $Q$ is the emission strength, and $g_\varphi$ is an advection--diffusion plume given $\varphi$.
For each of $360$ possible wind directions, the conditional posterior is Gaussian, so the exact posterior is a weighted mixture.
The scaffold instead conditions a single Gaussian with the prior's first two moments on the same readings (Appendix~\ref{app:plume}).
When several wind directions remain plausible, the posterior mean can be diffuse even though individual draws will always contain a narrow smoke plume.

\begin{figure}[t]
	\centering
	\includegraphics[width=\linewidth]{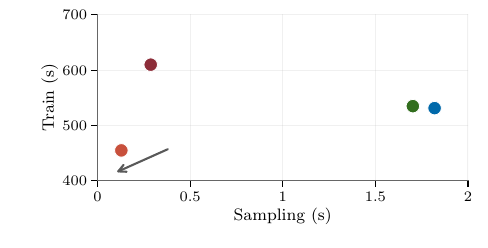}
	\caption{\dPlume{} training and sampling costs at $n=8192$, $d=1024$. Methods: \mGainMix{}~(\legmarkercirclefull{plumeGainMix}), \mGainDiff{}~(\legmarkercirclefull{plumeGainDiff}), \mFullDiff{}~(\legmarkercirclefull{plumeFullDiff}), \mDiff{}~(\legmarkercirclefull{plumeDiff}). The arrow indicates improvement; points show timing medians over three seeds. Training includes selection and excludes evaluation; sampling produces 100 draws for one context.}
	\label{fig:plume}
\end{figure}

\begin{figure}[t]
	\centering
	\includegraphics[width=\linewidth]{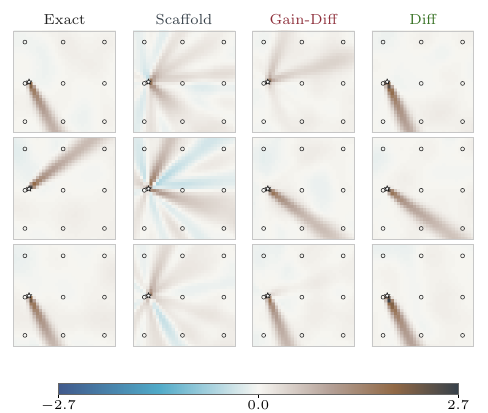}
	\caption{\dPlume{} posterior draws for one set of sensor observations (first three draws per method; rows are not paired). Exact draws are narrow plumes in different directions; the scaffold smears smoke across directions, while \mGainDiff{} and \mDiff{} recover localized plumes. Stars: source; circles: sensors.}
	\label{fig:plume-samples}
\end{figure}

The \texttt{Smoke} columns of \Cref{tab:main} report results at $n_\mathrm{train}=8192$.
The GP discrepancy baseline (\mKO{}) leaves the scaffold's CRPS essentially unchanged.
The presented corrections recover an increasing share of the predictive accuracy missing from the Gaussian scaffold.
Correcting the mean (\mScaffold{}+\mMean{}) closed $33\%$ of the scaffold-to-exact CRPS gap, and additional covariance calibration $\mRescale{}$ raises this to $42\%$.
Learning a distribution on the selected subspace yields a further substantial improvement: \mGainMix{} closes $85\%$ of the gap, and \mGainDiff{} reaches $90\%$ (\Cref{tab:main}, \texttt{Smoke} columns).
The selected diffusion correction achieves this using only $38$--$39$ of the $1024$ field coordinates.

The posterior samples illustrate the resulting improvement in plume structure (\Cref{fig:plume-samples}).
For the displayed readings, several wind directions remain plausible.
Exact draws contain individual narrow plumes in different directions, whereas scaffold draws spread smoke across several directions at once.
\mGainDiff{} recovers more localized plumes, although one of its displayed draws remains diffuse.
Across test contexts, its CRPS is close to that of the moment-matched Gaussian \mExactGP{} ($0.0457$ versus $0.0459$), while its PCD is about eight times lower ($154$ versus $1256$; \Cref{tab:main}).
Thus, similar marginal accuracy can coexist with substantially different sample plausibility.

\mFullDiff{} achieves slightly lower CRPS than \mGainDiff{}, although \mGainDiff{} has lower PCD at $n=8192$ (\Cref{tab:main}).
Direct conditional diffusion \mDiff{} achieves the lowest CRPS, Cram\'er distance and PCD among the learned methods.
The learning curves also favor the full-field models: both achieve lower mean CRPS than the selected corrections throughout the measured training sizes (Appendix~\ref{app:plume:curves}).
The contrast with \texttt{Synthetic} suggests that the useful inductive bias depends on the structure of the discrepancy: fixed bump directions favor learning in coefficient space, whereas uncertainty over coherent plume shapes may favor the spatial structure of the full-field U-Nets used here.
The selected correction, however, offers a clear tradeoff between predictive accuracy and reduced sampling cost (\Cref{fig:plume}).

\subsection{\texttt{CR3BP}}
\label{subsec:exp_cr3bp}

Finally, we examine whether a selected correction can improve numerical trajectory predictions.
The Gaussian scaffold is supplied by a probabilistic ODE solver~\citep{kramer2024implementing} for the planar circular restricted three-body problem~\citep{hairer1993solving,hennig2022probnum}.
The context is the initial state, and the target comprises positions and velocities at $501$ times ($d=2004$).
The dynamics are deterministic given the initial state, so the task is to correct numerical prediction error.
A high-accuracy numerical solve supplies the reference trajectory (Appendix~\ref{app:cr3bp}).

\begin{figure}[t]
 \centering
 \includegraphics[width=\linewidth]{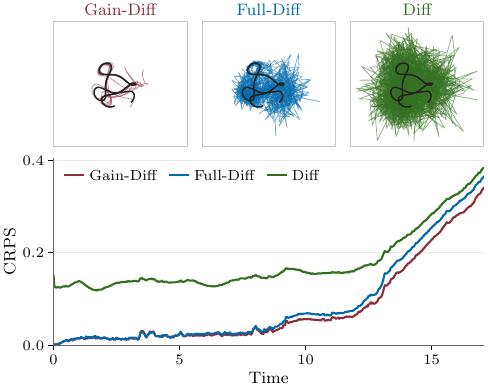}
 \caption{\dCRBP{} ($n=7500$, single seed). \textit{Top:} first five draws per method for one initial state, with the reference trajectory in black; \mGainDiff{}~\figlegendline{plumeGainDiff} stays close to the reference orbit, while \mFullDiff{}~\figlegendline{plumeFullDiff} and \mDiff{}~\figlegendline{plumeDiff} scatter. \textit{Bottom:} CRPS over time, averaged over the four state coordinates and all test contexts.}
 \label{fig:cr3bp-samples-crps}
\end{figure}

The \texttt{CR3BP} columns of \Cref{tab:main} report results at $n_\mathrm{train}=7500$.
As in \texttt{Smoke}, correcting the mean and covariance already improves substantially on the Gaussian scaffold.
Learning a distribution on the selected subspace improves accuracy further: \mGainDiff{} reaches a CRPS of $0.0815$, about half the solver's $0.1588$.
In contrast to \texttt{Smoke}, direct conditional diffusion \mDiff{} has higher CRPS than the scaffold.
For \mKO{}, likelihood fitting produces an over-dispersed prediction, where its log-score improves, but CRPS against the deterministic reference worsens (Appendix~\ref{app:cr3bp:results}).

The trajectory samples compare the three diffusion models (\Cref{fig:cr3bp-samples-crps} and \Cref{fig:cr3bp_pipeline}).
For the displayed initial state, \mGainDiff{} draws follow the reference orbit more closely, while \mFullDiff{} and direct conditional diffusion (\mDiff{}) produce visibly irregular paths.
Since PCD is unavailable for the point-mass reference, we assess physical consistency through Jacobi-constant drift (Appendix~\ref{app:cr3bp}), which measures changes in a state-derived quantity that should remain constant along each trajectory.
Both, \mGainDiff{} and \mGainMix{}, achieve lower field CRPS, energy scores and Jacobi errors than both full-field alternatives.

\section{Limitations and conclusion}
\label{sec:conclusion}

We developed a posterior-correction framework that uses a Gaussian reference to carry physical and observational information into a learned generative model.
Mean and covariance corrections address location and spread, while a richer probability law is learned on directions selected by estimated improvement over the calibrated reference.
The remaining directions retain Gaussian uncertainty.
This targeted correction recovers hidden modes in \texttt{Synthetic}, improves predictive accuracy and sample structure in \texttt{Smoke}, and reduces numerical trajectory error in \texttt{CR3BP}.

The framework leaves several sources of approximation error.
Even an exact law on the selected subspace leaves mismatch in the Gaussian complement and dependence between the two subspace unresolved.
Pooled calibration may miss context-dependent covariance structure, while gains estimated by the product-mixture selector may not carry over to the fitted diffusion model.
It remains open how to adapt global subspace selection when the leading directions of discrepancy vary strongly across contexts.
The selected rank depends on the data, resolution, and search budget.
In \texttt{CR3BP}, selection reaches the imposed rank cap, so the experiment does not establish an intrinsic correction dimension, but a useful regularization.
Its point-mass target also falls outside the theory's finite-KL assumptions.

\texttt{Smoke} illustrates that uncertainty in a single physical parameter can require coordinated corrections across many field directions.
The restricted correction improves predictive accuracy over the Gaussian reference and produces more plausible samples, while direct diffusion achieves higher predictive accuracy at a greater sampling cost.
\mGainDiff's subspace selection introduces additional fitting overhead, but reduces the dimensionality of the generative model and simplifies its training.

Together, these results suggest that a Gaussian approximation can reduce the scope of generative learning when its remaining discrepancy admits a compact correction.
Our framework makes this principle operational: use computation to construct a reference, use residual data to identify where a richer law adds value, and concentrate generative modeling there while retaining a Gaussian law on the complement.

\subsubsection*{Acknowledgements}

The authors gratefully acknowledge financial support by the European Research Council through ERC CoG Action 101123955 ANUBIS; the DFG Cluster of Excellence "Machine Learning - New Perspectives for Science", EXC 2064/1, project number 390727645; the DFG SPP 2298 (Project HE 7114/5-1), and the Carl Zeiss Foundation, (project "Certification and Foundations of Safe Machine Learning Systems in Healthcare"), as well as funds from the Ministry of Science, Research and Arts of the State of Baden-Württemberg.
The authors thank the International Max Planck Research School for Intelligent Systems (IMPRS-IS) for supporting Tobias Weber.

\printbibliography

\clearpage
\appendix
\thispagestyle{empty}
\onecolumn
\section*{Appendix}

\etocsettagdepth{mtoc}{none}
\etocsettagdepth{atoc}{subsection}
\etocsettocstyle{\section*{Contents}}{\bigskip}
\tableofcontents
\etocdepthtag.toc{atoc}

\section{Method details}
\label{app:method}

This section completes the decompositions of \Cref{sec:method} and then describes each stage as implemented.
\Cref{alg:app:fit} summarizes one full pipeline fit; per-problem values are collected in \Cref{tab:app:settings}.

\begin{figure}[H]
	\centering
	\includegraphics[width=\linewidth]{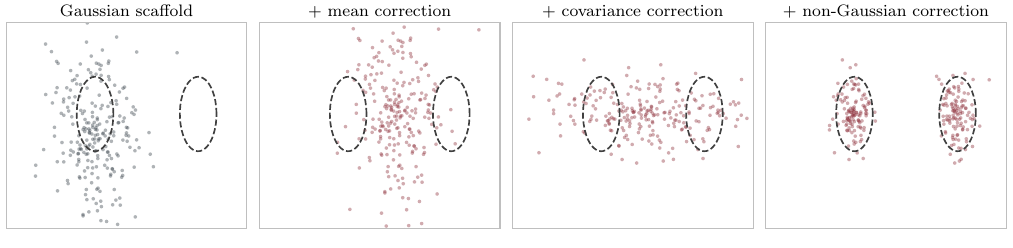}
	\caption{Illustration of the correction principle in two dimensions. Dashed contours mark two target modes. Correcting the mean and covariance accounts for location and spread, but leaves the Gaussian unable to represent both modes. Replacing the law along the horizontal direction recovers the bimodality while retaining a Gaussian in the vertical direction.}
	\label{fig:correction-principle}
\end{figure}

\subsection{Derivations}
\label{app:method:derivations}

Throughout, $P_c=\cL(f\mid c)$ has a density, finite second moments, mean $\mu_c$ and nonsingular covariance $\Sigma_c$, and all divergences are finite.
These assumptions concern the density identities; the deterministic CR3BP target does not satisfy them in ambient field space.
Its sample-based comparisons do not require finite KL divergence.
Statements are per context $c$ and the objectives of \Cref{sec:method} take their expectation over $c$.

\paragraph{Gaussian projection (\Cref{eq:kl_decomposition}).}
Let $G_c^\star=\cN(\mu_c,\Sigma_c)$ and $Q=\cN(m,C)$.
Since $\log G_c^\star-\log Q$ is a quadratic polynomial in $f$, its expectation depends on the law of $f$ only through the first two moments, which $P_c$ and $G_c^\star$ share. Hence
\begin{equation}
	\KL{P_c}{Q}-\KL{P_c}{G_c^\star}
	=\E_{P_c}\!\big[\log G_c^\star-\log Q\big]
	=\E_{G_c^\star}\!\big[\log G_c^\star-\log Q\big]
	=\KL{G_c^\star}{Q},
	\label{eq:app:pythagoras}
\end{equation}
and the closed form of the Gaussian divergence splits $\KL{G_c^\star}{Q}$ into the location and covariance terms of \Cref{eq:kl_decomposition}.

\paragraph{Mean step.}
For any shift $\delta$,
$\E_{P_c}\norm{f-m_c-\delta}^2_{C_c^{-1}}=\norm{\mu_c-m_c-\delta}^2_{C_c^{-1}}+\tr(C_c^{-1}\Sigma_c)$.
The precision-weighted loss is therefore twice the location term plus a constant in $\delta$, and it is minimized by $\delta=\mu_c-m_c$.

\paragraph{Covariance mismatch after pooled calibration.}
Covariance correction removes the pooled mismatch, but may leave a context-dependent contribution to the covariance term in the KL decomposition.
Assume the mean is exact, and let $u$ denote the whitened residual after exact pooled covariance calibration.
Write $\Sigma_c^u=\Cov(u\mid c)$, so that $\E_c\Sigma_c^u=I$.
The remaining expected covariance contribution to the Gaussian KL is
\begin{equation}
	\E_c\,\tfrac12\big[\tr\Sigma_c^u-\log\det\Sigma_c^u-d\big]
	=-\tfrac12\,\E_c\log\det\Sigma_c^u\;\ge\;0,
	\label{eq:app:heteroscedastic}
\end{equation}
where the equality uses $\E_c\Sigma_c^u=I$ and the inequality follows from concavity of $\log\det$.
For positive-definite covariances, this remainder vanishes if and only if $\Sigma_c^u=I$ for almost every context.
Any remaining context-dependent covariance mismatch must be addressed by the block law.

\paragraph{Block decomposition (\Cref{eq:kl_probabilistic_decomposition}).}
The map $u\mapsto(w,w_\perp)=(U\T u,U_\perp\T u)$ is orthogonal, so densities transform without a Jacobian.
Writing $\log p(w,w_\perp\mid c)=\log p(w\mid c)+\log p(w_\perp\mid c)+\log\frac{p(w,w_\perp\mid c)}{p(w\mid c)\,p(w_\perp\mid c)}$ inside $\KL{P_{u\mid c}}{q(\cdot\mid c)}$ gives the block term, the complement term and the mutual information $I(w;w_\perp\mid c)$.

\paragraph{Gain (\Cref{eq:gain}).}
Subtracting the decomposition for $q$ from the same decomposition for the calibrated Gaussian $\cN(0,I_d)$ cancels the complement and the mutual-information terms:
\begin{align}
	\E_c\KL{P_{u\mid c}}{\cN(0,I_d)}-\E_c\KL{P_{u\mid c}}{q(\cdot\mid c)}
	 & =\E_c\!\big[\KL{P_{w\mid c}}{\cN(0,I_k)}-\KL{P_{w\mid c}}{q_\theta(\cdot\mid c)}\big] \nonumber \\
	 & =\cG(U,q_\theta).
	\label{eq:app:gain}
\end{align}
The gain is thus the reduction of total divergence achieved by that block law up to finite-sampling errors.
It is bounded by $\E_c\KL{P_{w\mid c}}{\cN(0,I_k)}$, which is attained at $q_\theta=P_{w\mid c}$.

\paragraph{Gain under a factorized selection density.}
Selection uses a product of one-dimensional laws, $q(w\mid c)=\prod_j q_j(w_j\mid c)$.
With the total correlation $\mathrm{TC}(w\mid c)=\KL{P_{w\mid c}}{\prod_jP_{w_j\mid c}}\ge0$,
\begin{equation}
	\KL{P_{w\mid c}}{\cN(0,I_k)}=\sum_{j=1}^k\KL{P_{w_j\mid c}}{\cN(0,1)}+\mathrm{TC}(w\mid c).
	\label{eq:app:product}
\end{equation}
The gain a product contrast can measure is therefore a lower bound on the gain available to a joint law such as the block DDPM; the difference is the dependence inside the block.

\paragraph{Resolution $\tau$.}
Selection scores the residual and the Gaussian reference after both are convolved with $\cN(0,\tau^2I)$.
For a fixed pair, the data-processing inequality~\citep{cover2006elements} guarantees that the smoothed divergence never exceeds the original one, and discrepancies whose amplitude in the units of $f$ is small against $\tau$ contribute little to it.

\subsection{Covariance mismatch correction}
\label{app:method:rescale}

\paragraph{Coordinates.}
Let $r=f-\hat\mu_c$ be the mean-corrected residual.
A \emph{shared-covariance scaffold} uses the same covariance $C_c=C$ for every context (\dSynth{}, \dPlume{}); a \emph{context-dependent-covariance scaffold} uses a covariance $C_c$ that varies with $c$ (\dCRBP{}).
In our shared-covariance examples, $C$ is numerically low rank, so we add an isotropic variance floor and whiten as $z=(C+\sigma^2I)^{-1/2}r$.
For \dSynth{}, we choose $\sigma$ by grid search on the training Gaussian log score; for \dPlume{}, we set $\sigma=\tau$.
For \dCRBP{}, we use the context-specific factor $C_c=S_cS_c\T$ and whiten as $z=S_c^{-1}r$, without adding a floor.

\paragraph{Rescaling with a sampling-noise threshold.}
Let $\hat\lambda_i$ be the variance of the whitened residuals $z$ along the $i$th eigenvector of their sample covariance, and $\tilde\lambda_i$ the variance the calibrated reference assigns to that direction.
Even when the scaffold covariance is correct, the $\hat\lambda_i$ of $n$ residuals in $m$ dimensions spread over the Marchenko--Pastur interval $[(1-\sqrt{m/n})^2,(1+\sqrt{m/n})^2]$~\citep{marchenko1967distribution}.
We therefore set $\tilde\lambda_i=\hat\lambda_i$ only if $\kappa(\hat\lambda_i)$ exceeds its value at both edges, and $\tilde\lambda_i=1$ otherwise.
For \dCRBP{}, this test runs on the full field space.
For shared scaffolds, it runs on the range of $C$ and on the complement's principal components whose eigenvalues exceed the Marchenko--Pastur edge of the others, debiased following~\citet{baik2006eigenvalues}; all other complement directions share their average sample variance.
The resulting variances define $L_c$ and $u=L_c^{-1}r$.

\subsection{Subspace selection}
\label{app:method:selection}

\paragraph{Contrast.}
The selection density is a product of one-dimensional Gaussian mixtures with $K=3$ components per coordinate.
Its parameters are shared across contexts in \dSynth{} and predicted from the context by an MLP in \dPlume{} and \dCRBP{}.
Each coordinate's held-out gain is $g_j=\frac{1}{n_{\mathrm{val}}}\sum_{\mathrm{val}}\big[\log q_j(w_j\mid c)-\log\cN(w_j;0,1)\big]$.

\paragraph{Resolution.}
The resolution is $\tau=c_\tau\cdot\mathrm{RMS}(r)$ with $c_\tau\in\{0,0.2\}$ per problem (\Cref{tab:app:settings}).
Training and validation residuals receive $\tau\xi$, $\xi\sim\cN(0,I)$, and the calibration is refitted on the smoothed residuals with floor $\tau$ (shared) or covariance $C_c+\tau^2I$ (per context).
After selection, the directions are mapped back to the unsmoothed coordinates.

\paragraph{Proposals and search.}
The gain is a non-convex function of the frame $U$, so gradient ascent from an arbitrary start can stall in a poor local optimum.
We therefore start from three proposals, each restricted to the resolved calibration directions:
\begin{itemize}\setlength{\itemsep}{1pt}
	\item \emph{Residual PCA}: principal components of $r$ in function space, mapped into $u$. They capture the residual structure with the largest amplitude in the units of $f$.
	\item \emph{FastICA}~\citep{hyvarinen1999fast}: independent components of $u$, ranked by held-out gain. FastICA seeks maximally non-Gaussian projections, which is the structure the gain rewards.
	\item \emph{$\hat\lambda$-span}: after recalibration, the at most $64$ axes with the largest $\hat\lambda_i>1$, rotated within their span by the same Stiefel gradient ascent as the refinement below, using a pooled mixture contrast. Separated modes widen the residual along the directions that separate them, so they tend to appear as excess variance.
\end{itemize}
The leading $64$ directions of each proposal form a pool, scored by $\sum_j\max(g_j,0)$ under a fitted contrast (see above).
The top $k_{\max}=64$ directions of the best pool are refined jointly with the contrast by gradient ascent on the Stiefel manifold of orthonormal frames, through a smooth parametrization~\citep{edelman1998geometry,lezcano2019trivializations}.
If refinement lowers the gain, the pool is kept.
For per-context scaffolds, the refinement runs inside a $64$-dimensional span assembled from the proposals and is then polished in the full resolved space.

\paragraph{Rank selection.}
A coordinate is kept if $g_j>\max\{0.01,(3K-1)/n_{\mathrm{val}}\}$, where $3K-1$ is the parameter count of a one-dimensional mixture ($0.016$ nats at $n_{\mathrm{val}}=500$).
After mapping back to $\tau=0$, a freshly fitted contrast drops coordinates with $g_j\le0$.
The \emph{law guard} falls back to the Gaussian when a mixture block has no positive held-out gain over $\cN(0,I_k)$ or, on \dSynth{}, when a block DDPM's best validation state is its initialization.
These decisions use the validation split, never the test contexts; because the same split early-stops the contrast, the reported gains are optimistic and the threshold does not control an error rate.
An empty block means that the search found no gain, not that the residual is Gaussian.

\begin{algorithm}[t]
	\caption{Fitting the corrected posterior for one training seed.}
	\label{alg:app:fit}
	\begin{algorithmic}[1]
		\Require training and validation pairs $(f,c)$, scaffold $(m_c,C_c)$, resolution factor $c_\tau$
		\State Fit the mean shift $\delta_\phi$; set $\hat\mu_c=m_c+\delta_\phi(c)$ and $r=f-\hat\mu_c$
		\State Calibrate the covariance, giving $L_c$ and $u=L_c^{-1}r$ \Comment{\Cref{app:method:rescale}}
		\State Set $\tau=c_\tau\,\mathrm{RMS}(r)$, add $\cN(0,\tau^2I)$ noise to $r$ and recalibrate
		\For{each proposal (PCA, FastICA, $\hat\lambda$-span)}
		\State Fit the contrast on its leading $64$ directions and score them by $\sum_j\max(g_j,0)$
		\EndFor
		\State Refine the best pool on the Stiefel manifold; keep the pool if refinement lowers the score
		\State Keep the directions whose $g_j$ exceeds the threshold
		\State Map them back to $\tau=0$ and drop those without held-out gain there, giving $U\in\R^{d\times k}$
		\If{$k>0$}
		\State Fit the block law $q_\theta$ (mixture or DDPM) on $w=U\T u$ and apply the law guard
		\EndIf
		\State \Return the sampler of \Cref{eq:posterior_sampling}, which is \mRescale{} when $k=0$
	\end{algorithmic}
\end{algorithm}

\subsection{Block laws and sampling}
\label{app:method:laws}

\paragraph{Block laws.}
All block laws are fitted on $w=U\T u$ at $\tau=0$.
\mGainMix{} refits the factorized Gaussian mixture contrast from scratch.
\mGainDiff{} uses a DDPM~\citep{ho2020denoising} with $T=200$ steps, a cosine noise schedule~\citep{nichol2021improved} and $v$-prediction~\citep{salimans2022progressive}.
Its MLP denoiser is conditioned on $c$, and ancestral sampling clips the predicted $\hat w_0$ to the training range widened on each side by half its width.
Coordinates are standardized per dimension.
On \dSynth{}, whose readings carry no information about the bump signs, both block laws are unconditional and the DDPM predicts $\epsilon$.

\paragraph{Full-residual baseline.}
\mFullDiff{} takes $k=d$ \emph{without} the covariance calibration: it models the mean-corrected residual $r$ (shared scaffolds) or $S_c^{-1}r$ (per-context scaffolds) directly.

\paragraph{Corrected draw.}
A draw is $f^\star=\hat\mu_c+L_c\big(Uw+(I-UU\T)\epsilon\big)$ with $w\sim q_\theta(\cdot\mid c)$ and $\epsilon\sim\cN(0,I_d)$; this has the law of \Cref{eq:posterior_sampling} without forming $U_\perp$.
Directions outside the resolved calibration directions are never selected, so they always remain Gaussian.

\FloatBarrier
\section{Shared experimental setup}
\label{app:setup}

All problems share the method names, the training protocol and the metrics defined here.
Problem-specific choices are collected in \Cref{tab:app:settings} and explained in the problem sections.

\subsection{Methods}
\label{app:setup:roster}

\Cref{tab:app:roster} defines every method by its center, its Gaussian part and its learned law.
The structured residual corrections share the anchor mean stage $\hat\mu_c=m_c+\delta_\phi(c)$ of their problem.
The mean of a learned block law can produce a further shift, so this anchor is not necessarily the predictive mean.
Block and full-residual laws are unconditional on \dSynth{} and conditional on the other problems.

\begin{table}[h]
	\centering
	\scriptsize
	\setlength{\tabcolsep}{4pt}
	\caption{
		Method roster.
		$L_c$: calibrated factor (\Cref{app:method:rescale}); $k$: block size; mixtures are products of one-dimensional Gaussian mixtures; \mKO{} is shown before conditioning on readings.
		Grey rows require the exact posterior.
	}
	\label{tab:app:roster}
	\begin{tabular}{@{}llllll@{}}
		\toprule
		Method                        & Anchor                                                 & Gaussian part                      & Learned law                        & Problems                           & Defined in                                \\
		\midrule
		\mScaffold{}                  & $m_c$                                                  & $C_c$                              & --                                 & all                                & \Cref{sec:method}                         \\
		\mScaffold{}+\mMean{}         & $\hat\mu_c$                                            & $C_c$                              & --                                 & all                                & \Cref{sec:method}                         \\
		\mRescale{}                   & $\hat\mu_c$                                            & $L_cL_c\T$                         & --                                 & all                                & \Cref{app:method:rescale}                 \\
		\mGainMix{}                   & $\hat\mu_c$                                            & $L_c(I-UU\T)L_c\T$                 & mixture on $k$ coordinates         & all                                & \Cref{app:method:laws}                    \\
		\mGainDiff{}                  & $\hat\mu_c$                                            & $L_c(I-UU\T)L_c\T$                 & DDPM on $k$ coordinates            & all                                & \Cref{app:method:laws}                    \\
		\mFullDiff{}                  & $\hat\mu_c$                                            & --                                 & DDPM on all $d$ coordinates        & all                                & \Cref{app:method:laws}                    \\
		\mKO{}                        & $\rho_m m_0(c)+\mu$                                    & $\rho^2C_0(c)+K_\delta$            & --                                 & all                                & below                                     \\
		\mDiff{}                      & --                                                     & --                                 & conditional DDPM on $f$            & all                                & below                                     \\
		\midrule
		\textcolor{gray}{\mExactGP{}} & \textcolor{gray}{$\mu_c$}                              & \textcolor{gray}{$\Sigma_c$}       & \textcolor{gray}{--}               & \textcolor{gray}{all but \dCRBP{}} & \textcolor{gray}{\Cref{sec:introduction}} \\
		\textcolor{gray}{\mExact{}}   & \multicolumn{3}{l}{\textcolor{gray}{draws from $P_c$}} & \textcolor{gray}{all but \dCRBP{}} & \textcolor{gray}{problem sections}                                                                                  \\
		\bottomrule
	\end{tabular}
\end{table}

\paragraph{\mKO{}.}
Following \citet{kennedy2001bayesian}, the field is $f=\rho_m m_0(c)+\rho\,\eta+\delta$, where $\eta\sim\cN(0,C_0(c))$ is the scaffold's model before conditioning on the readings, $\delta$ is a Gaussian-process discrepancy with a mean $\mu$ per grid point and covariance $K_\delta=s^2(R_\ell+\nu I)$, and the readings are $y=Hf+e$ with $e\sim\cN(0,\lambda I)$.
The mean scale $\rho_m$ is chosen by generalized least squares, $(\rho,s^2,\ell,\nu)$ are fitted by type-II maximum likelihood (L-BFGS-B within boxes) on the training and validation sets, and $\lambda$ is the average of $(y-Hf)^2$ over the training readings, i.e.\ the empirical reading-noise variance.
With $\rho_m=\rho=1$ and $\delta=0$, \mKO{} is the scaffold.
Its discrepancy prior is shared across contexts; conditioning can still change the resulting mean with the readings, but it does not learn an arbitrary context-dependent discrepancy process.

\paragraph{\mDiff{}.}
A conditional DDPM on the whole field that does not use the scaffold ($v$-prediction, $T=200$, the cosine schedule and clipping rule of \Cref{app:method:laws}).
It is an MLP on \dSynth{} and \dCRBP{}, and a U-Net with the context as input channels on \dPlume{}.
Its RMSE uses the mean of its draws.

\subsection{Training protocol}
\label{app:setup:protocol}

\paragraph{Stages and data.}
The stages are fitted in the order mean, calibration, selection, law, each on the same training split and frozen before the next; there is no joint fine-tuning.
The validation split early-stops every network and supplies the held-out gains.

\paragraph{Optimization.}
The learned neural components use AdamW~\citep{loshchilov2019decoupled} with cosine learning-rate schedules and gradient clipping; diffusion models keep exponential moving averages of their weights (decay $0.999$)~\citep{song2020improved}.
Budgets and learning rates differ by problem and component, as summarized in \Cref{tab:app:settings}.
Diffusion validation averages four fixed-noise evaluations; the checkpoint with the lowest validation objective is restored.
Validation frequency and patience are set per component.
Diffusion models standardize their coordinates per dimension (U-Nets: per-pixel centering, one shared scale), and networks standardize the context per column.

\paragraph{Seeds and splits.}
Each final comparison runs three seeds.
A seed draws the data (on \dCRBP{}, re-splits one pool), initializes the networks and seeds the evaluation.
Smaller training sets are subsets of larger ones, and the validation and test sets do not depend on $n_{\mathrm{train}}$.
Architectures and training settings were chosen during development using exploratory sweeps and then fixed for the reported comparisons.

\paragraph{Cost reporting.}
Fit-plus-evaluation wall times use the last completed run per method and seed.
The timings exclude prior dataset and scaffold preparation and do not separate sampling from evaluation.
For \dPlume{} we report training and sampling separately.

\begin{table}[t]
	\centering
	\scriptsize
	\setlength{\tabcolsep}{3pt}
	\caption{
		Per-problem settings.
		Learning rates are for the block DDPM and \mDiff{}; budgets are in optimizer steps.
	}
	\label{tab:app:settings}
	\resizebox{\textwidth}{!}{%
		\begin{tabular}{@{}llll@{}}
			\toprule
			                                          & \dSynth{}                       & \dPlume{}                        & \dCRBP{}                             \\
			\midrule
			$d$                                       & $512$                           & $1024$ ($32\times32$)            & $2004$ ($501\times4$)                \\
			$n_{\mathrm{train}}$ / $n_{\mathrm{val}}$ & $2000$ / $500$                  & $8192$ / $500$                   & $7500$ / $500$                       \\
			Mean                                      & linear (ridge)                  & MLP $2\times128$, span-projected & MLP $2\times256$                     \\
			Floor $\sigma$                            & log-score grid                  & $\tau$                           & -- (per context)                     \\
			$c_\tau$ / contrast                       & $0$ / pooled                    & $0.2$ / conditional              & $0.2$ / conditional                  \\
			Search span                               & resolved                        & resolved                         & $64$                                 \\
			Mixture law                               & product, unconditional          & product, conditional             & product, conditional                 \\
			Block DDPM                                & MLP $256$, $\epsilon$, uncond.  & MLP $512\times4$, $v$            & MLP $256$, $v$                       \\
			\quad batch / budget / lr                 & $64$ / $100$k / $2\cdot10^{-3}$ & $64$ / $30$k / $2\cdot10^{-3}$   & $64$ / $100$k / $2\cdot10^{-3}$      \\
			\mFullDiff{}                              & MLP $1024$, uncond.             & U-Net                            & MLP $4008$                           \\
			\mDiff{}                                  & MLP $1024$                      & U-Net                            & MLP $4008$                           \\
			\quad budget / lr                         & $100$k / $10^{-3}$              & $16{,}384$ / $2\cdot10^{-4}$     & $100$k / $10^{-3}$                   \\
			\mKO{} discrepancy                        & periodic SE + nugget            & SE (ARD) + nugget                & SE$(t)\otimes B_{4\times4}$ + nugget \\
			\bottomrule
		\end{tabular}%
	}
\end{table}

\subsection{Evaluation and metrics}
\label{app:setup:metrics}

\paragraph{Protocol.}
Each method draws $E=100$ samples for each of $500$ test contexts, with random numbers shared across methods.
Where exact draws are available, \mExact{} and the reference distances use separate ensembles; the distance between the two gives the Monte Carlo floor.
Marginal scores are averaged over coordinates and then contexts, and joint scores over contexts.
Tables report the mean $\pm$ standard deviation over seeds; these describe seed variation and are not confidence intervals.

\paragraph{Notation.}
For one context, $X,X'$ are independent draws of a method, $Z,Z'$ independent exact draws, $f^\star$ the truth and $j=1,\dots,d$ a coordinate.

\paragraph{CRPS.}
The continuous ranked probability score~\citep{gneiting2007strictly}, averaged over coordinates,
\begin{equation}
	\mathrm{CRPS}=\frac1d\sum_{j=1}^d\Big[\E|X_j-f^\star_j|-\tfrac12\E|X_j-X'_j|\Big],
	\label{eq:setup:crps}
\end{equation}
estimated with the fair ensemble estimator (distinct members in the second term).
$\mathrm{CRPS}_{\text{view}}$ restricts the average to a subset of coordinates, such as the sensors or a region.

\paragraph{Cram\'er distance to the exact posterior.}
\begin{equation}
	d_{\mathrm C}=\frac1d\sum_{j=1}^d\Big[2\E|X_j-Z_j|-\E|X_j-X'_j|-\E|Z_j-Z'_j|\Big],
	\label{eq:setup:cramer}
\end{equation}
which is twice the Cram\'er distance $\int(F_j-G_j)^2$ between the marginal distribution functions~\citep{szekely2013energy}.
It is estimated without bias and can therefore be slightly negative when the two laws agree.

\paragraph{Energy score.}
$\mathrm{ES}=\E\norm{X-f^\star}-\tfrac12\E\norm{X-X'}$~\citep{gneiting2008assessing}, with the fair estimator, scores the joint field against the truth.

\paragraph{Coverage.}
With $E$ draws $X^{(1)},\dots,X^{(E)}$ and the rank $r_j=\sum_{e=1}^E\mathds{1}\{X^{(e)}_j<f^\star_j\}$ of the truth in coordinate $j$,
\begin{equation}
	C_{90}=\frac1d\sum_{j=1}^d\Big|\,[r_j,\,r_j+1]\cap\big[0.05(E+1),\,0.95(E+1)\big]\Big|,
	\label{eq:setup:coverage}
\end{equation}
averaged over contexts, where $|\cdot|$ is the length of an interval.
For a calibrated ensemble each rank has probability $1/(E+1)$, and the partial credit at the two boundary ranks makes $C_{90}=0.90$ exactly for any $E$.

\paragraph{Physical consistency.}
Pointwise accuracy does not ensure that a complete draw has plausible joint structure.
For example, a smoke field may combine several wind directions that are individually plausible but incompatible within one realization.
We define the physical consistency distance (PCD) to compare how generated and exact draws score under the exact posterior.
We compare score distributions because minimizing the average negative log density alone can reward concentration at high-density points rather than reproduce posterior uncertainty, an intuition related to typicality tests~\citep{nalisnick2019detecting}.

For context $c$, we assign each complete draw the score
\begin{equation}
	s_c(f)=-\log p_{\tau_{\mathrm{eval}}}(f\mid c),
	\qquad
	p_\tau(\cdot\mid c)
	=p(\cdot\mid c)*\cN(0,\tau^2I),
\end{equation}
where $p(\cdot\mid c)$ is the exact posterior.
For independent method draws $X,X'$ and exact draws $Z,Z'$, the distance for this context is
\begin{equation}
	\begin{aligned}
		\mathrm{PCD}_c
		={}&2\E|s_c(X)-s_c(Z)|\\
		&-\E|s_c(X)-s_c(X')|
		-\E|s_c(Z)-s_c(Z')|.
	\end{aligned}
	\label{eq:setup:pcd}
\end{equation}
This is the Cram\'er distance of~\Cref{eq:setup:cramer} applied to scalar scores rather than individual field coordinates.
We compute it from the generated and independent exact reference ensembles, excluding identical-member pairs in the within-ensemble terms, and average over test contexts.

Smoothing prevents tiny deviations in directions with negligible posterior variance from dominating the score.
We set $\tau_{\mathrm{eval}}=0.2\,\mathrm{RMS}(f-m_c)$ over the full training pool and field coordinates, using the canonical scaffold mean $m_c$.
This scale is shared across methods and training-set sizes within each problem and seed; smoothing applies only to the scoring density, without perturbing the evaluated draws.

Lower PCD indicates closer agreement between the score distributions.
The population distance is zero for exact draws; its unbiased ensemble estimate may be slightly negative.
PCD can reveal joint-structure errors and excessive concentration that alter these scores, but matching score distributions does not imply matching posterior distributions.
Its scale depends on the problem and smoothing choice, so numerical values are compared within each problem.

\paragraph{Gap closed.}
The share of the scaffold-to-exact CRPS gap a method closes is
\begin{equation}
	\frac{\mathrm{CRPS}_{\mathrm{scaffold}}-\mathrm{CRPS}_{\mathrm{method}}}{\mathrm{CRPS}_{\mathrm{scaffold}}-\mathrm{CRPS}_{\mathrm{exact}}},
	\label{eq:setup:gap}
\end{equation}
which is $0\%$ for \mScaffold{} and $100\%$ for \mExact{}.

\paragraph{RMSE.}
The root-mean-square error of a point prediction $\hat f$,
\begin{equation}
	\mathrm{RMSE}=\Big(\E_c\,\frac1d\sum_{j=1}^d\big(\hat f_j-f^\star_j\big)^2\Big)^{1/2},
	\label{eq:setup:rmse}
\end{equation}
where $\hat f$ is the anchor $\hat\mu_c$ for the corrected methods and the predictive mean otherwise.

\FloatBarrier
\section{Dataset: \dSynth{}}
\label{app:msgp}

In \dSynth{} the scaffold misses Gaussian structure and non-Gaussian bump modes.
The readings carry no information about the non-Gaussian part, so the missing structure lies in the same directions in every context.

\subsection{Data, exact posterior and scaffold}
\label{app:msgp:setup}

\paragraph{Field and readings.}
On the periodic grid $x_i=i/d$, $d=512$, the field is $f=g+h+Bz$ with independent periodic GPs $g$ and $h$, kernel
\begin{equation}
	k_{v,\ell}(x,x')=v\exp\!\big(-2\sin^2(\pi(x-x'))/\ell\big),
	\label{eq:msgp:kernel}
\end{equation}
with $v=1$, $\ell=0.341$ for the coarse component $g$ and $v=0.3$, $\ell=0.083$ for the fine component $h$.
Sixteen sensors sit at $x=0,\Delta,\dots,15\Delta$ with $\Delta=1/16$ and read $y_i=f(x_{s_i})+\sigma_i\varepsilon_i$, alternately precise ($\sigma_i=0.02$) and noisy ($\sigma_i=0.15$).
Each column of $B$ is a bump $a\sin^2(\pi\,(x-x_{s_j})/\Delta)$ on the $j$-th gap between neighbouring sensors and zero elsewhere, so $AB=0$ for the sensor operator $A$.
The signs $z\in\{-1,+1\}^{16}$ are uniform, and the amplitude $a$ places the two modes along each unit bump at about $\pm3$ posterior standard deviations.

\paragraph{Exact posterior.}
Let $(m_G(y),C_G)$ be the posterior of $g+h$ under the true sensor noise.
Then
\begin{equation}
	p(f\mid y)=2^{-16}\sum_{z}\cN\big(m_G(y)+Bz,\;C_G\big),
	\label{eq:msgp:exact}
\end{equation}
and an exact draw is a Gaussian draw plus fresh uniform signs.
\mExactGP{} is $\cN(m_G(y),C_G+BB\T)$.

\paragraph{Scaffold.}
The scaffold is the GP posterior of $g$ alone, conditioned on all readings with one pooled noise level $\bar\sigma=\mathrm{RMS}_i(\sigma_i)=0.107$.
Its mean is affine in $y$, its covariance is shared across contexts and has rank $35$.
It misses $h$ (which shifts the conditional mean and adds variance mostly outside the range of $C_c$), the unequal sensor precision and the bumps.

\subsection{Methods and training}
\label{app:msgp:methods}

Because the exact mean is affine in $y$, the mean stage is a ridge regression ($10^{-6}$) of the residual on the readings.
The floor is chosen on the log-score grid ($\sigma=0.32$), and the calibration covers the $35$ range directions.
Selection uses the pooled contrast at $\tau=0$, since the bump law does not depend on the readings and both block laws are unconditional.
\mFullDiff{} and \mDiff{} are MLPs of width $1024$; \mDiff{} standardizes the field per grid point.

\subsection{Evaluation}
\label{app:msgp:eval}

The physical consistency score is the negative log density of the convolved exact mixture~\eqref{eq:msgp:exact} at $\tau=0.2\cdot\mathrm{RMS}(f-m_c)\approx0.08$.

\subsection{Additional results}
\label{app:msgp:results}

The main table uses $n_{\mathrm{train}}=2000$, with all reported methods complete.
\Cref{tab:synthetic-appendix} adds seed spread, the energy score and RMSE to the main-table metrics.

\begin{table}[t]
\centering\normalsize
\setlength{\tabcolsep}{2pt}
\caption{\dSynth{} ($d=512$, $n=2000$): mean $\pm$ standard deviation over three seeds (500 contexts $\times$ 100 draws). Corrected methods share the RMSE of their mean stage; $k$ is the block size.}
\label{tab:synthetic-appendix}
\begin{tabular*}{\linewidth}{@{\extracolsep{\fill}}lrrrrrrr@{}}
\toprule
Method & CRPS & $d_{\mathrm C}$ & $C_{90}$ & PCD & ES & RMSE & $k$ \\
\midrule
\mScaffold{} & 0.2859{\tiny$\pm$0.0016} & 0.1884{\tiny$\pm$0.0006} & 0.30{\tiny$\pm$0.00} & 667{\tiny$\pm$5} & 7.86{\tiny$\pm$0.04} & 0.4036{\tiny$\pm$0.0016} & -- \\
\mKO{} & 0.2130{\tiny$\pm$0.0011} & 0.0427{\tiny$\pm$0.0002} & 0.90{\tiny$\pm$0.00} & 1242{\tiny$\pm$5} & 6.46{\tiny$\pm$0.03} & 0.4055{\tiny$\pm$0.0017} & -- \\
\mDiff{} & 0.2208{\tiny$\pm$0.0013} & 0.0575{\tiny$\pm$0.0012} & 0.86{\tiny$\pm$0.00} & 1256{\tiny$\pm$18} & 6.79{\tiny$\pm$0.04} & 0.4241{\tiny$\pm$0.0024} & -- \\
\addlinespace[2pt]
\hdashline[0.4pt/2pt]
\addlinespace[2pt]
\mScaffold{}+\mMean{} & 0.2744{\tiny$\pm$0.0012} & 0.1659{\tiny$\pm$0.0002} & 0.33{\tiny$\pm$0.00} & 512{\tiny$\pm$2} & 7.69{\tiny$\pm$0.03} & 0.3955{\tiny$\pm$0.0012} & 0 \\
\mRescale{} & 0.2014{\tiny$\pm$0.0007} & 0.0193{\tiny$\pm$0.0002} & 0.94{\tiny$\pm$0.00} & 406{\tiny$\pm$0} & 6.26{\tiny$\pm$0.02} & 0.3955{\tiny$\pm$0.0012} & 0 \\
\mGainMix{} & 0.1936{\tiny$\pm$0.0007} & 0.0026{\tiny$\pm$0.0001} & 0.89{\tiny$\pm$0.00} & 23{\tiny$\pm$3} & 6.27{\tiny$\pm$0.02} & 0.3955{\tiny$\pm$0.0012} & 16 \\
\mGainDiff{} & 0.1934{\tiny$\pm$0.0005} & 0.0027{\tiny$\pm$0.0002} & 0.89{\tiny$\pm$0.00} & 32{\tiny$\pm$3} & 6.27{\tiny$\pm$0.02} & 0.3955{\tiny$\pm$0.0012} & 16 \\
\mFullDiff{} & 0.1947{\tiny$\pm$0.0006} & 0.0050{\tiny$\pm$0.0003} & 0.87{\tiny$\pm$0.01} & 958{\tiny$\pm$241} & 6.28{\tiny$\pm$0.02} & 0.3955{\tiny$\pm$0.0012} & 512 \\
\midrule
\mExactGP{} & 0.2007{\tiny$\pm$0.0009} & 0.0175{\tiny$\pm$0.0000} & 0.95{\tiny$\pm$0.00} & 407{\tiny$\pm$1} & 6.24{\tiny$\pm$0.02} & 0.3940{\tiny$\pm$0.0014} & -- \\
\mExact{} & 0.1921{\tiny$\pm$0.0007} & 0.0000{\tiny$\pm$0.0002} & 0.90{\tiny$\pm$0.00} & 0{\tiny$\pm$0} & 6.24{\tiny$\pm$0.02} & 0.3940{\tiny$\pm$0.0014} & -- \\
\bottomrule
\end{tabular*}
\end{table}

\begin{figure}[t]
	\centering
	\includegraphics[width=\linewidth]{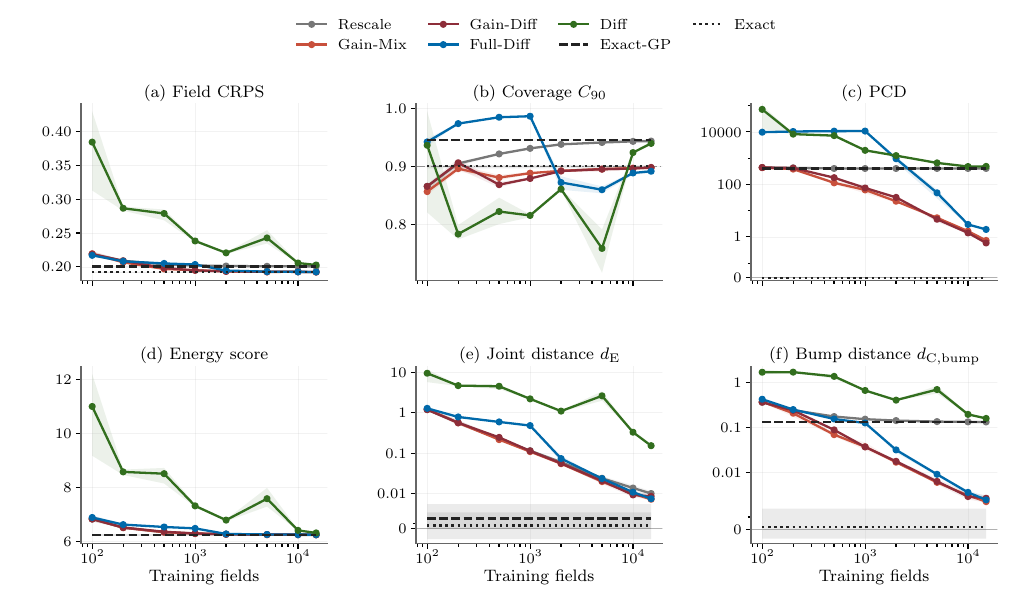}
	\caption{\dSynth{} learning curves for six further metrics, using 500 test contexts and 100 draws per context. Lines show means and bands span three seeds. Dashed and dotted lines show \mExactGP{} and \mExact{}, respectively; the horizontal rule in the coverage panel marks $0.90$. Distance panels use logarithmic scales with a linear region near zero.}
	\label{fig:synthetic-metrics}
\end{figure}

\paragraph{Learning curve.}
\label{app:msgp:curves}
The curve in \Cref{fig:synthetic_curve} uses $n_{\mathrm{train}}\in\{100,200,500,1000,2000,5000,10{,}000,15{,}000\}$, with three seeds at every point for every plotted method.
The main table compares methods at $n=2000$; the larger training sizes show how the gain methods and full-residual diffusion approach the exact reference as more data become available.
\Cref{fig:synthetic-metrics} shows the complementary marginal, joint and physical-consistency metrics across the same training sizes.

\begin{figure}[!htbp]
	\centering
	\includegraphics[width=\linewidth]{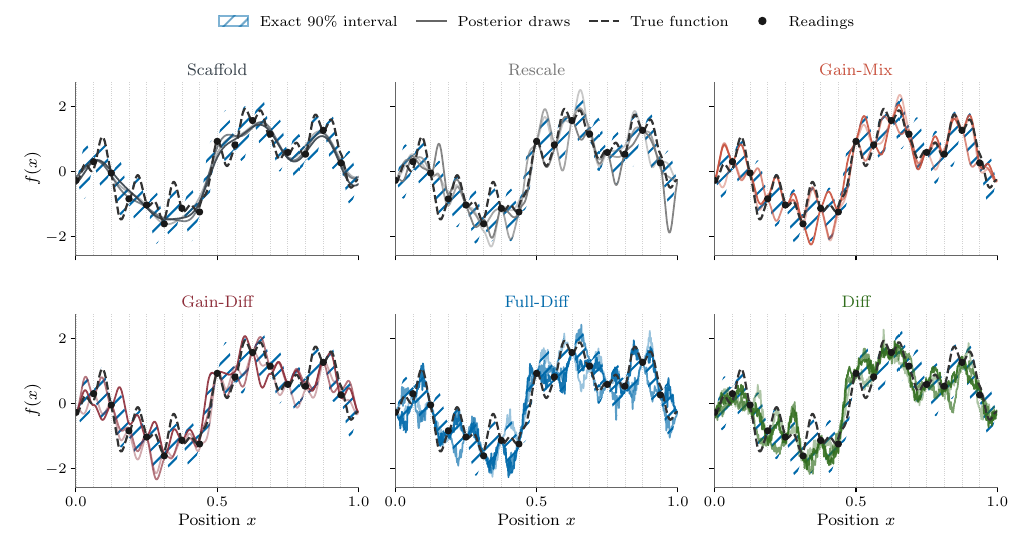}
	\caption{\dSynth{} posterior draws at $n=2000$ (seed 10): three draws per method for the first test context, over the true function (dashed) and the readings (dots). Hatching marks the exact pointwise 90\% intervals and dotted lines the sensors. The readings do not determine the bump signs, so valid draws need not match the true function between sensors.}
	\label{fig:synthetic-function-overlay}
\end{figure}

\begin{figure}[!htbp]
	\centering
	\includegraphics[width=\linewidth]{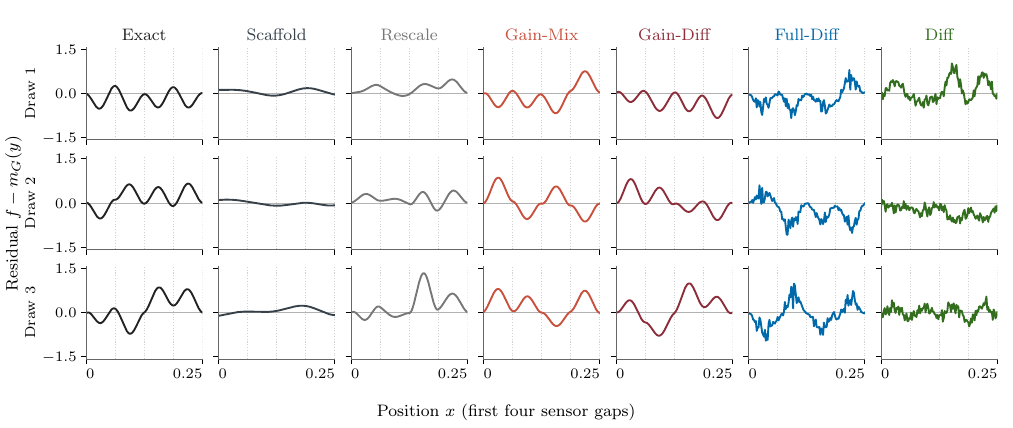}
	\caption{The draws of \Cref{fig:synthetic-function-overlay} and three exact draws on the first four sensor gaps, centered on the exact posterior mean $m_G(y)$; dotted lines mark the sensors.}
	\label{fig:synthetic-samples}
\end{figure}

\paragraph{Selected subspace.}
From $n_{\mathrm{train}}=500$ on, the selector keeps $k=16$ coordinates on every seed, the number of bumps.
With fewer fields fewer coordinates clear the bar, and at $n_{\mathrm{train}}=100$ the subspace selection falls back to the Gaussian.
\paragraph{Posterior samples.}
\Cref{fig:synthetic-function-overlay} shows three draws per method for the first test context at $n=2000$, on the full domain over the generating function, the sensor readings and the exact posterior's pointwise 90\% intervals.
\Cref{fig:synthetic-samples} zooms into the first four sensor gaps, adds exact draws and centers all draws on the exact posterior mean to expose the structure between sensors.

\paragraph{Cost.}
\label{app:msgp:cost}
\Cref{tab:msgp-cost} reports combined fit-plus-evaluation wall time at $n=2000$.
At this scale, learning fewer stochastic coordinates is therefore not equivalent to reducing the full computational cost.

\begin{table}[t]
\centering\normalsize
\setlength{\tabcolsep}{6pt}
\caption{\dSynth{}, $n=2000$: median fit-plus-evaluation time (s) over three seeds on an RTX 2080 Ti. Includes selection; excludes data preparation.}
\label{tab:msgp-cost}
\begin{tabular}{@{}lr@{}}
\toprule
Method & Fit + evaluation (s) \\
\midrule
\mRescale{} & 40 \\
\mGainMix{} & 349 \\
\mGainDiff{} & 361 \\
\mFullDiff{} & 276 \\
\mDiff{} & 208 \\
\bottomrule
\end{tabular}
\end{table}

\FloatBarrier
\section{Dataset: \dPlume{}}
\label{app:plume}

In \dPlume{} nine readings of a smoke plume constrain an unknown wind direction, sharply in some contexts and only to several sectors in others.
The exact posterior is a mixture over the wind direction, the scaffold a single Gaussian with the prior's first two moments, so the missing structure depends on the readings.

\subsection{Data, exact posterior and scaffold}
\label{app:plume:setup}

\paragraph{Field and readings.}
A point source at $x_s=(0.15,0.5)$ releases a tracer into a steady wind of speed $u=1$ and direction $\varphi$, with diffusivity $\kappa=0.003$.
The concentration is the free-space Green's function of $u\cdot\nabla c=\kappa\Delta c+\delta(x-x_s)$,
\begin{equation}
	g_\varphi(x)\propto e^{a\cos\theta}K_0(a),\qquad a=\frac{u\norm{x-x_s}}{2\kappa},
	\label{eq:plume:green}
\end{equation}
with $\theta$ the angle between $x-x_s$ and the wind and $K_0$ the modified Bessel function.
We normalize $g_\varphi$ to one on its axis at distance $0.4$ and cap it smoothly near the source, $g\leftarrow3\tanh(g/3)$; the result is a narrow ray ($1/e$ half-width $0.08$ at distance $0.5$).
The field on the $32\times32$ cell centers is
\begin{equation}
	f=b+Q\,g_\varphi,\qquad b\sim\mathcal{GP}(0,K),\quad Q\sim\cN(q_0,s_Q^2),\quad p(\varphi_k)\propto e^{\cos\varphi_k},
	\label{eq:plume:field}
\end{equation}
with a squared-exponential $K$ (standard deviation $0.05$, lengthscale $0.15$), $q_0=1$, $s_Q=0.1$, and $\varphi$ on a grid of $360$ angles, so the posterior below is exact rather than a quadrature.
The wind blows away from the sensors' side ($|\varphi|>90^\circ$) with prior probability $0.22$.
The $3\times3$ sensors on $[0.1,0.9]^2$ are snapped to the nearest cell centers and read $y=Hf+\varepsilon$ with $\sigma=0.02$.

\paragraph{Exact posterior.}
Given $\varphi=\varphi_k$, the field is linear in $(b,Q)$ and hence Gaussian, and so are the readings: $y\mid\varphi_k\sim\cN(q_0h_k,S_k)$ with $h_k=Hg_k$ and $S_k=HKH\T+\sigma^2I+s_Q^2h_kh_k\T$.
Therefore
\begin{equation}
	p(f\mid y)=\sum_{k=1}^{360}w_k(y)\,\cN\big(m_k(y),\,C_k-C_kH\T S_k^{-1}HC_k\big),\qquad w_k(y)\propto p(\varphi_k)\,\cN(y;q_0h_k,S_k),
	\label{eq:plume:exact}
\end{equation}
with $C_k=K+s_Q^2g_kg_k\T$ and $m_k=q_0g_k+C_kH\T S_k^{-1}(y-q_0h_k)$; weights, means and pathwise draws are available in closed form.

\paragraph{Scaffold.}
The scaffold conditions the Gaussian with the prior's mean and covariance on $y$,
\begin{equation}
	m=q_0\bar g,\qquad P=K+(q_0^2+s_Q^2)\textstyle\sum_kp_kg_kg_k\T-q_0^2\bar g\bar g\T,\qquad\bar g=\sum_kp_kg_k.
	\label{eq:plume:scaffold}
\end{equation}
The prior moments determine the best linear prediction from the readings and its residual covariance.
They do not determine the true conditional mean or covariance at each context, or a context-dependent choice of wind sector; scaffold draws can superpose several rays.

\begin{figure*}[t]
	\centering
	\includegraphics[width=\textwidth]{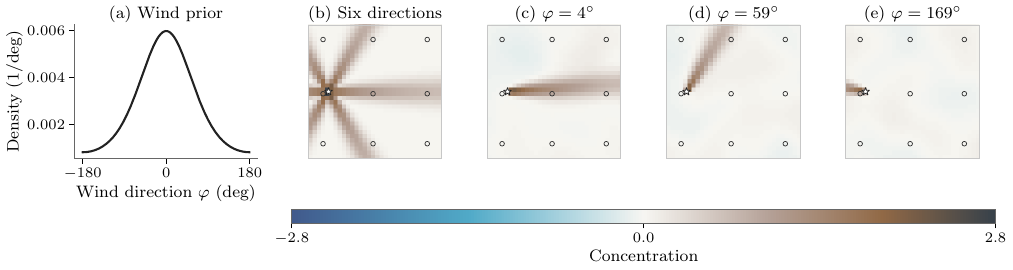}
	\caption{
		\dPlume{}.
		(a) Prior over the wind direction.
		(b) Smoke plume shapes $g_\varphi$ for six directions, with the source ($\star$) and the sensors ($\circ$).
		(c--e) Three draws of the field: a sharp ray whose direction changes from draw to draw, on a quiet background.
	}
	\label{fig:plume-problem}
\end{figure*}

\subsection{Methods and training}
\label{app:plume:methods}

The mean stage is an MLP ($2\times128$) of the observations.
We choose $\tau=0.2\cdot\mathrm{RMS}(r)\approx0.026$, and the calibration covers $309$ range directions of $C_c$ and $197$ complement spikes.
Selection uses the conditional contrast density with an MLP encoder of the nine readings.
\mFullDiff{} and \mDiff{} share one U-Net (channels $32,64,128,128$, two blocks per level, attention at the coarsest level, cf. \citet{ho2020denoising}); \mDiff{} receives the readings as a value map and a mask and through an MLP that modulates every block.
\mKO{} collapses to the scaffold. Its fitted $\rho\approx1$ and the discrepancy variance sits at its lower bound.

\subsection{Evaluation}
\label{app:plume:eval}

The physical consistency score is the negative log density of the exact mixture~\eqref{eq:plume:exact} convolved with $\cN(0,\tau^2I)$, with $Q$ integrated out, at $\tau=0.2\cdot0.163\approx0.033$.
Besides the field we report the views \emph{source} (the disk of radius $0.2$ around the source), \emph{upwind} ($x_1<0.15$) and \emph{downwind} (the rest).

\subsection{Additional results}
\label{app:plume:results}

\Cref{tab:plume-appendix} reports the retained methods on the final round in the order of \Cref{tab:main}.

\begin{table*}[t]
\centering
\scriptsize
\setlength{\tabcolsep}{2pt}
\caption{
	\dPlume{}, reported methods: field and source-disk scores, 500 test contexts $\times$ 100 draws, mean $\pm$ std over seeds 10, 11, 12 ($n_{\mathrm{train}} = 8192$).
	$d_{\mathrm{C}}$: Cramér distance to the exact posterior (Exact: its split-half floor); PCD: the Cramér distance between posterior log-density score distributions for generated and exact draws; RMSE: root mean squared error of the predictive mean; $k$: the selected block's size. Best non-reference value in bold, references in grey.
}
\label{tab:plume-appendix}
\resizebox{\textwidth}{!}{%
\begin{tabular}{@{}lcccccccccc@{}}
\toprule
 & \multicolumn{6}{c}{field} & \multicolumn{3}{c}{source} & \multicolumn{1}{c}{} \\
Method & CRPS & $d_{\mathrm{C}}$ & Cov90 & PCD & ES & RMSE & CRPS & $d_{\mathrm{C}}$ & Cov90 & $k$ \\
\midrule
\mScaffold{} & 0.0714{\tiny$\pm$0.0002} & 0.0585{\tiny$\pm$0.0001} & 0.92 & 7072.3{\tiny$\pm$10.9} & 3.561{\tiny$\pm$0.012} & 0.1631{\tiny$\pm$0.0005} & 0.1342{\tiny$\pm$0.0020} & 0.1306{\tiny$\pm$0.0013} & 0.93 & -- \\
\mKO{} & 0.0714{\tiny$\pm$0.0002} & 0.0584{\tiny$\pm$0.0001} & 0.92 & 7078.7{\tiny$\pm$18.5} & 3.560{\tiny$\pm$0.012} & 0.1631{\tiny$\pm$0.0006} & 0.1342{\tiny$\pm$0.0021} & 0.1307{\tiny$\pm$0.0014} & 0.93 & -- \\
\mDiff{} & \textbf{0.0434{\tiny$\pm$0.0009}} & \textbf{0.0014{\tiny$\pm$0.0002}} & 0.89 & \textbf{36.5{\tiny$\pm$1.3}} & \textbf{2.515{\tiny$\pm$0.029}} & \textbf{0.1353{\tiny$\pm$0.0017}} & \textbf{0.0714{\tiny$\pm$0.0015}} & \textbf{0.0028{\tiny$\pm$0.0004}} & \textbf{0.89} & -- \\
\addlinespace[2pt]
\hdashline[0.4pt/2pt]
\addlinespace[2pt]
\mScaffold{}+\mMean{} & 0.0619{\tiny$\pm$0.0005} & 0.0385{\tiny$\pm$0.0001} & 0.94 & 6906.3{\tiny$\pm$38.3} & 2.966{\tiny$\pm$0.017} & 0.1372{\tiny$\pm$0.0012} & 0.1135{\tiny$\pm$0.0020} & 0.0882{\tiny$\pm$0.0006} & 0.94 & 0 \\
\mRescale{} & 0.0595{\tiny$\pm$0.0006} & 0.0336{\tiny$\pm$0.0001} & 0.92 & 5190.5{\tiny$\pm$35.1} & 2.914{\tiny$\pm$0.018} & 0.1372{\tiny$\pm$0.0012} & 0.1082{\tiny$\pm$0.0020} & 0.0775{\tiny$\pm$0.0006} & 0.92 & 0 \\
\mGainMix{} & 0.0473{\tiny$\pm$0.0010} & 0.0088{\tiny$\pm$0.0009} & 0.91 & 1022.4{\tiny$\pm$33.8} & 2.667{\tiny$\pm$0.057} & 0.1372{\tiny$\pm$0.0012} & 0.0806{\tiny$\pm$0.0024} & 0.0216{\tiny$\pm$0.0016} & 0.92{\tiny$\pm$0.01} & 38{\tiny$\pm$1} \\
\mGainDiff{} & 0.0457{\tiny$\pm$0.0012} & 0.0051{\tiny$\pm$0.0009} & \textbf{0.90} & 153.9{\tiny$\pm$15.0} & 2.650{\tiny$\pm$0.068} & 0.1372{\tiny$\pm$0.0012} & 0.0762{\tiny$\pm$0.0022} & 0.0115{\tiny$\pm$0.0014} & \textbf{0.91{\tiny$\pm$0.01}} & 38{\tiny$\pm$1} \\
\mFullDiff{} & 0.0450{\tiny$\pm$0.0011} & 0.0049{\tiny$\pm$0.0005} & 0.88 & 248.7{\tiny$\pm$25.6} & 2.567{\tiny$\pm$0.040} & 0.1372{\tiny$\pm$0.0012} & 0.0755{\tiny$\pm$0.0019} & 0.0117{\tiny$\pm$0.0007} & 0.88 & 1024 \\
\midrule
\textcolor{gray}{\mExactGP{}} & \textcolor{gray}{0.0459{\tiny$\pm$0.0007}} & \textcolor{gray}{0.0062{\tiny$\pm$0.0001}} & \textcolor{gray}{0.91} & \textcolor{gray}{1256.2{\tiny$\pm$44.7}} & \textcolor{gray}{2.569{\tiny$\pm$0.024}} & \textcolor{gray}{0.1339{\tiny$\pm$0.0017}} & \textcolor{gray}{0.0778{\tiny$\pm$0.0014}} & \textcolor{gray}{0.0157{\tiny$\pm$0.0006}} & \textcolor{gray}{0.91} & -- \\
\textcolor{gray}{\mExact{}} & \textcolor{gray}{0.0429{\tiny$\pm$0.0007}} & \textcolor{gray}{0.0000{\tiny$\pm$0.0001}} & \textcolor{gray}{0.90} & \textcolor{gray}{0.0} & \textcolor{gray}{2.485{\tiny$\pm$0.024}} & \textcolor{gray}{0.1339{\tiny$\pm$0.0017}} & \textcolor{gray}{0.0703{\tiny$\pm$0.0012}} & \textcolor{gray}{-0.0000{\tiny$\pm$0.0001}} & \textcolor{gray}{0.90} & -- \\
\bottomrule
\end{tabular}%
}
\end{table*}

\begin{figure}[!htbp]
	\centering
	\includegraphics[width=\linewidth]{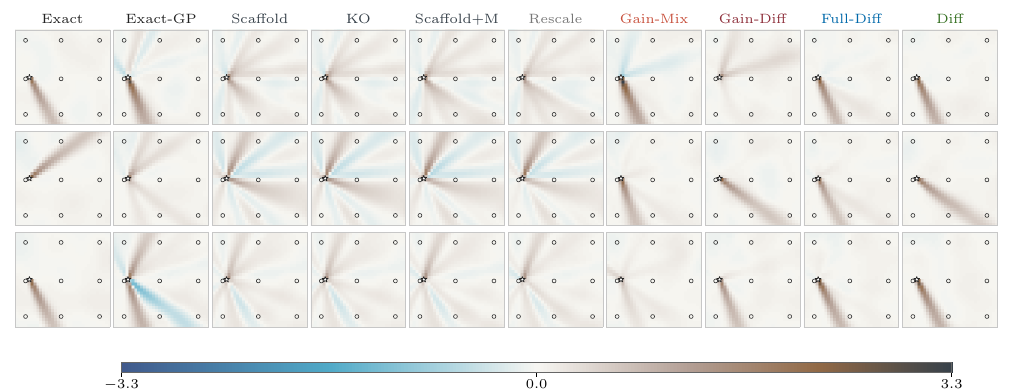}
	\caption{\dPlume{} posterior draws for all retained methods and the two exact references. The readings, training size ($n=8192$), training seed ($10$) and first three draws match \Cref{fig:plume-samples}; rows are not paired realizations. \mExactGP{} matches the exact posterior's conditional mean and covariance.
		All panels share a concentration scale: brown is positive, blue is negative. Stars mark the source; circles mark sensors.}
	\label{fig:plume-samples-extended}
\end{figure}

\paragraph{Posterior samples.}
\Cref{fig:plume-samples-extended} extends \Cref{fig:plume-samples} to all retained corrections and baselines, alongside the exact posterior and its moment-matched Gaussian.

\paragraph{Learning curve.}
\label{app:plume:curves}
The \dPlume{} U-Nets keep a fixed optimizer-step budget as the training set changes.
The block methods require $n\ge2\cdot309=618$ fields for calibration.

\begin{figure}[t]
	\centering
	\includegraphics[width=\linewidth]{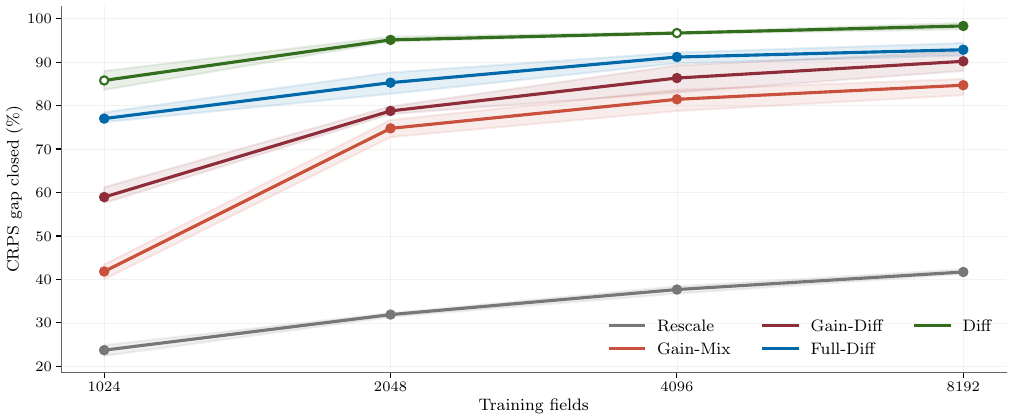}
	\caption{\dPlume{} learning curve: share of the scaffold-to-exact CRPS gap closed. Lines show seed means and bands span seeds. Filled points have three seeds; open points have two (scratch diffusion at $n=1024$ and $4096$). The curve starts at $n=1024$, where all methods have results. Scratch diffusion leads throughout the measured range.}
	\label{fig:plume-learning}
\end{figure}

The \dPlume{} curve shows improvement with more training data for each residual correction, but no crossover in favor of the gain methods.
The difference between \mGainDiff{} and direct diffusion narrows with increasing data; the direct U-Net still leads at $n=8192$.
This supports reporting the restricted model as a practical alternative, without a general claim that scientific scaffolds improve sample efficiency.

\paragraph{Cost.}
\label{app:plume:cost}
\Cref{tab:plume-cost} reports combined fit-plus-evaluation wall time at $n=8192$.
The gain methods have lower combined times than the full-field U-Nets, but the separate timings show that this is not a fitting advantage.
For \dPlume{}, \Cref{fig:plume} separates training from sampling.

\begin{table}[t]
\centering\normalsize
\setlength{\tabcolsep}{6pt}
\caption{\dPlume{}, $n=8192$: median fit-plus-evaluation time (s) over three seeds on an L40S. Includes selection; excludes data preparation.}
\label{tab:plume-cost}
\begin{tabular}{@{}lr@{}}
\toprule
Method & Fit + evaluation (s) \\
\midrule
\mRescale{} & 80 \\
\mGainMix{} & 527 \\
\mGainDiff{} & 687 \\
\mFullDiff{} & 1270 \\
\mDiff{} & 2983 \\
\bottomrule
\end{tabular}
\end{table}

\FloatBarrier
\section{Dataset: \texttt{CR3BP}}
\label{app:cr3bp}

In \dCRBP{} (Circular Restricted Three-Body Problem) the scaffold is a probabilistic ODE solver's Gaussian prediction for trajectories of the planar circular restricted three-body problem~\citep[Sec.~40.3]{hennig2022probnum}.
A trajectory is a deterministic function of its initial state, so the reference law is a point mass at the truth.
We score predictions against a high-accuracy reference trajectory and omit \mExact{}, \mExactGP{} and the redundant marginal distance $d_{\mathrm C}$.
The KL identities in \Cref{sec:method} assume a target density with finite divergences and do not apply to this point-mass reference.

\subsection{Data, reference trajectory and scaffold}
\label{app:cr3bp:setup}

\paragraph{Dynamics.}
In rotating coordinates with state $z=(x,y,v_x,v_y)$, mass parameter $\mu$, $\mu_1=1-\mu$, $r_1=\sqrt{(x+\mu)^2+y^2}$ and $r_2=\sqrt{(x-\mu_1)^2+y^2}$,
\begin{equation}
	\dot x=v_x,\quad\dot y=v_y,\quad
	\dot v_x=x+2v_y-\mu_1\frac{x+\mu}{r_1^3}-\mu\frac{x-\mu_1}{r_2^3},\quad
	\dot v_y=y-2v_x-\mu_1\frac{y}{r_1^3}-\mu\frac{y}{r_2^3},
	\label{eq:cr3bp:rhs}
\end{equation}
with $\mu=0.012277471$ and the Arenstorf orbit $z_A=(0.994,0,0,-2.0015851063790825)$ of period $T=17.0652$~\citep{hairer1993solving}.

\paragraph{Initial states and data.}
An initial state is the Arenstorf state propagated to a random phase and perturbed, $z_0=\Phi_{\varphi}(z_A)+\eta$, with $\varphi=(c_j+U[-0.05,0.05])\,T$, $c_j\in\{0,0.25,0.5,0.75\}$, and $\eta\sim\cN\big(0,\operatorname{diag}(5\cdot10^{-4},5\cdot10^{-4},10^{-3},10^{-3})^2\big)$.
The reference trajectory is integrated with \texttt{DOP853} (relative and absolute tolerances $10^{-12}$ and $10^{-13}$) over one period and stored at $501$ equispaced times, with four state coordinates per time, so $d=2004$.
Initial states are rejected if a coarse RK45 solve exceeds RMSE $1.5$, RMS Jacobi-constant drift $1.0$ or radius $2.0$ (relative and absolute tolerances $3\cdot10^{-3}$ and $3\cdot10^{-6}$).
One pool of $10{,}400$ trajectories is re-split per seed into $500$ test, $500$ validation and $9400$ training trajectories, of which the runs use $7500$.

\paragraph{Scaffold.}
The scaffold is a probabilistic ODE filter~\citep{kramer2024implementing} with a twice-integrated Wiener prior, first-order Taylor linearization, dynamic calibration, adaptive steps and smoothing, from which $32$ pathwise draws are interpolated to the grid.
Its covariance is $C_c=\alpha^2DD\T+\sigma^2I$, with $D$ the draws centered at the solver mean and scaled by $1/\sqrt{31}$, $\alpha=1$ and $\sigma=0.25$, fixed after a development sweep of held-out log-likelihood.
The experiment tests corrections to this floored solver prediction.
The context is $c=z_0$.

\subsection{Methods and training}
\label{app:cr3bp:methods}

The mean stage is an MLP ($2\times256$) of $z_0$ trained by mean squared error on trajectory residuals.
Covariance calibration uses the full pooled basis with the context-dependent scaffold, without adding a further floor.
Selection uses $\tau=0.2\cdot\mathrm{RMS}(r)\approx0.059$ and a conditional contrast on $z_0$, with at most $64$ directions in the search span.
The block DDPM is an MLP conditioned on $z_0$; \mFullDiff{} and \mDiff{} are MLPs of width $4008$.
\mKO{} uses the solver's Gaussian without the added floor and the discrepancy $K_\delta=R_\ell(t)\otimes B+\nu I$, where $R_\ell$ is squared-exponential and $B$ is a full $4\times4$ channel covariance.
Its likelihood is fitted on the resolved span of the pooled reference trajectories and solver means.

\subsection{Evaluation}
\label{app:cr3bp:eval}

CRPS, ES and $C_{90}$ are scored against the reference trajectory; for a point-mass reference, the marginal distance of \Cref{eq:setup:cramer} is twice CRPS.
The views are the positions, the velocities and the final state.
In place of PCD we use the Jacobi constant $C_J=x^2+y^2+2\big(\mu_1/r_1+\mu/r_2\big)-(v_x^2+v_y^2)$, which the dynamics conserve.
A draw's score is its drift $s(f)=\mathrm{RMS}_t\big[C_J(f_t)-C_J(z_0)\big]$ ($2\cdot10^{-12}$ for the truth).
The \emph{Jacobi error} is the median over contexts and draws of $|s(X)-s(f^\star)|$, reducing sensitivity to large drift values.

\subsection{Additional results}
\label{app:cr3bp:results}

\Cref{tab:cr3bp-appendix} reports the retained methods in the order of \Cref{tab:main}.

\begin{table}[t]
\centering
\normalsize
\setlength{\tabcolsep}{2pt}
\caption{\dCRBP{} ($d=2004$, $n=7500$): mean $\pm$ std over seeds 10--12 (500 contexts $\times$ 100 draws). Jacobi error is defined in \Cref{app:cr3bp:eval}; $k$ is the block size. Best mean in bold (coverage closest to $0.90$).}
\label{tab:cr3bp-appendix}
\begin{tabular*}{\linewidth}{@{\extracolsep{\fill}}lrrrrrrrr@{}}
\toprule
Method & CRPS & ES & Jacobi err. & C$_{90}$ & CRPS$_{\mathrm{pos}}$ & CRPS$_{\mathrm{vel}}$ & CRPS$_{\mathrm{final}}$ & $k$ \\
\midrule
\mScaffold{} & 0.1588{\tiny$\pm$0.0002} & 11.26{\tiny$\pm$0.05} & 1.46{\tiny$\pm$0.01} & \textbf{0.89} & 0.1713{\tiny$\pm$0.0020} & 0.1462{\tiny$\pm$0.0023} & 0.592{\tiny$\pm$0.003} & -- \\
\mKO{} & 0.2726{\tiny$\pm$0.0108} & 19.03{\tiny$\pm$0.59} & 5.50{\tiny$\pm$0.26} & 0.96 & 0.2910{\tiny$\pm$0.0119} & 0.2541{\tiny$\pm$0.0097} & 1.091{\tiny$\pm$0.025} & -- \\
\mDiff{} & 0.1785{\tiny$\pm$0.0036} & 10.91{\tiny$\pm$0.27} & 2.89{\tiny$\pm$0.01} & 0.99 & 0.2031{\tiny$\pm$0.0038} & 0.1538{\tiny$\pm$0.0034} & 0.384{\tiny$\pm$0.014} & -- \\
\addlinespace[2pt]
\hdashline[0.4pt/2pt]
\addlinespace[2pt]
\mScaffold{}+\mMean{} & 0.1178{\tiny$\pm$0.0028} & 7.83{\tiny$\pm$0.26} & 2.01{\tiny$\pm$0.03} & 0.93 & 0.1217{\tiny$\pm$0.0023} & 0.1138{\tiny$\pm$0.0036} & 0.397{\tiny$\pm$0.015} & 0 \\
\mRescale{} & 0.0916{\tiny$\pm$0.0024} & 7.60{\tiny$\pm$0.23} & 0.79{\tiny$\pm$0.02} & 0.94{\tiny$\pm$0.01} & 0.0911{\tiny$\pm$0.0015} & 0.0921{\tiny$\pm$0.0035} & 0.376{\tiny$\pm$0.011} & 0 \\
\mGainMix{} & 0.0844{\tiny$\pm$0.0028} & 7.24{\tiny$\pm$0.26} & 0.68{\tiny$\pm$0.02} & 0.96 & 0.0821{\tiny$\pm$0.0025} & 0.0866{\tiny$\pm$0.0033} & 0.372{\tiny$\pm$0.011} & 64 \\
\mGainDiff{} & \textbf{0.0815{\tiny$\pm$0.0031}} & \textbf{7.06{\tiny$\pm$0.34}} & \textbf{0.44{\tiny$\pm$0.03}} & 0.93 & \textbf{0.0799{\tiny$\pm$0.0023}} & \textbf{0.0830{\tiny$\pm$0.0043}} & \textbf{0.355{\tiny$\pm$0.016}} & 64 \\
\mFullDiff{} & 0.0902{\tiny$\pm$0.0024} & 7.65{\tiny$\pm$0.22} & 1.26{\tiny$\pm$0.01} & 0.97 & 0.0897{\tiny$\pm$0.0016} & 0.0907{\tiny$\pm$0.0033} & 0.374{\tiny$\pm$0.009} & 2004 \\
\bottomrule
\end{tabular*}
\end{table}

\paragraph{Cost.}
\label{app:cr3bp:cost}
\Cref{tab:cr3bp-cost} reports combined fit-plus-evaluation wall time at $n=7500$; separate sampling timings are not available.

\begin{table}[t]
\centering\normalsize
\setlength{\tabcolsep}{6pt}
\caption{\dCRBP{}, $n=7500$: median fit-plus-evaluation time (s) over three seeds on an RTX 2080 Ti. Includes selection; excludes data preparation.}
\label{tab:cr3bp-cost}
\begin{tabular}{@{}lr@{}}
\toprule
Method & Fit + evaluation (s) \\
\midrule
\mRescale{} & 275 \\
\mGainMix{} & 1081 \\
\mGainDiff{} & 1362 \\
\mFullDiff{} & 875 \\
\mDiff{} & 932 \\
\bottomrule
\end{tabular}
\end{table}

\FloatBarrier

\end{document}